\documentclass[10pt,twocolumn,letterpaper]{article}

\usepackage{times}
\usepackage{epsfig}
\usepackage{graphicx}
\usepackage{amsmath}
\usepackage{amssymb}
\usepackage{booktabs}
\usepackage{multirow}
\usepackage{bbding}
\usepackage{pifont}

\usepackage{marvosym}
\usepackage[most]{tcolorbox} 
\newcommand{\draco}{{\em DraCo}}

\usepackage[pagenumbers]{cvpr} 

\definecolor{cvprblue}{rgb}{0.21,0.49,0.74}
\usepackage[pagebackref,breaklinks,colorlinks,allcolors=cvprblue]{hyperref}

\def\paperID{} 
\def\confName{CVPR}
\def\confYear{2026}

\title{\includegraphics[height=1.5em]{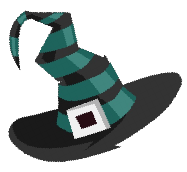} \draco: Draft as CoT for Text-to-Image Preview and Rare Concept Generation}

\author{Dongzhi Jiang$^{*1}$, Renrui Zhang$^{*\dagger 1}$\textsuperscript{\Letter}, Haodong Li$^{4}$, Zhuofan Zong$^{1}$, Ziyu Guo$^{2}$, Jun He$^{3}$, Claire Guo$^{5}$, \\Junyan Ye$^{3}$, Rongyao Fang$^{1}$, Weijia Li$^{3}$, Rui Liu$^{1}$\textsuperscript{\Letter}, Hongsheng Li$^{1}$\textsuperscript{\Letter}\vspace{0.3cm}\\
  $^1$CUHK MMLab\quad
  $^2$CUHK IMIXR\quad
  $^3$Sun Yat-Sen University\vspace{0.1cm}\quad
  $^4$SCUT\vspace{0.1cm}\quad
  $^5$CUHK (Shenzhen)\vspace{0.1cm}\\
  \texttt{\{dzjiang, renruizhang\}@link.cuhk.edu.hk}\\
  \small $^*$Equal Contribution\hspace{0.4cm} $^\dagger$Project Leader\hspace{0.4cm}  \textsuperscript{\Letter}Corresponding author
}

\begin{document}
\maketitle
\begin{abstract}
Recent unified multimodal large language models (MLLMs) have shown impressive capabilities, incorporating chain-of-thought (CoT) reasoning for enhanced text-to-image generation. However, existing approaches remain limited, either treating the model merely as a standalone generator or relying on abstract textual planning. 
To this end, we propose \textbf{Draft-as-CoT (DraCo)}, a novel interleaved reasoning paradigm that fully leverages both textual and visual contents in CoT for better planning and verification.
Our method first generates a low-resolution draft image as preview, providing more concrete and structural visual planning and guidance.
Then, we employ the model's inherent understanding capability to verify potential semantic misalignments between the draft and input prompt, and performs refinement through selective corrections with super-resolution. 
In this way, our approach addresses two fundamental challenges: 
the coarse-grained nature of textual planning and the difficulty in generating rare attribute combinations.
To support training, we curate DraCo-240K, aiming to enhance three atomic capabilities spanning general correction, instance manipulation, and layout reorganization.
Supported by DraCo-CFG, a specialized classifier-free guidance (CFG) strategy for interleaved reasoning, DraCo achieves a tremendous increase on GenEval (+8\%), Imagine-Bench (+0.91), and GenEval++ (+3\%), significantly outperforming direct generation and other generation methods empowered by CoT.
The project is at \url{https://github.com/CaraJ7/DraCo}.
\end{abstract}    
\section{Introduction}
\label{sec:intro}

\begin{figure}[tp]
    \centering
      \includegraphics[width=\linewidth]{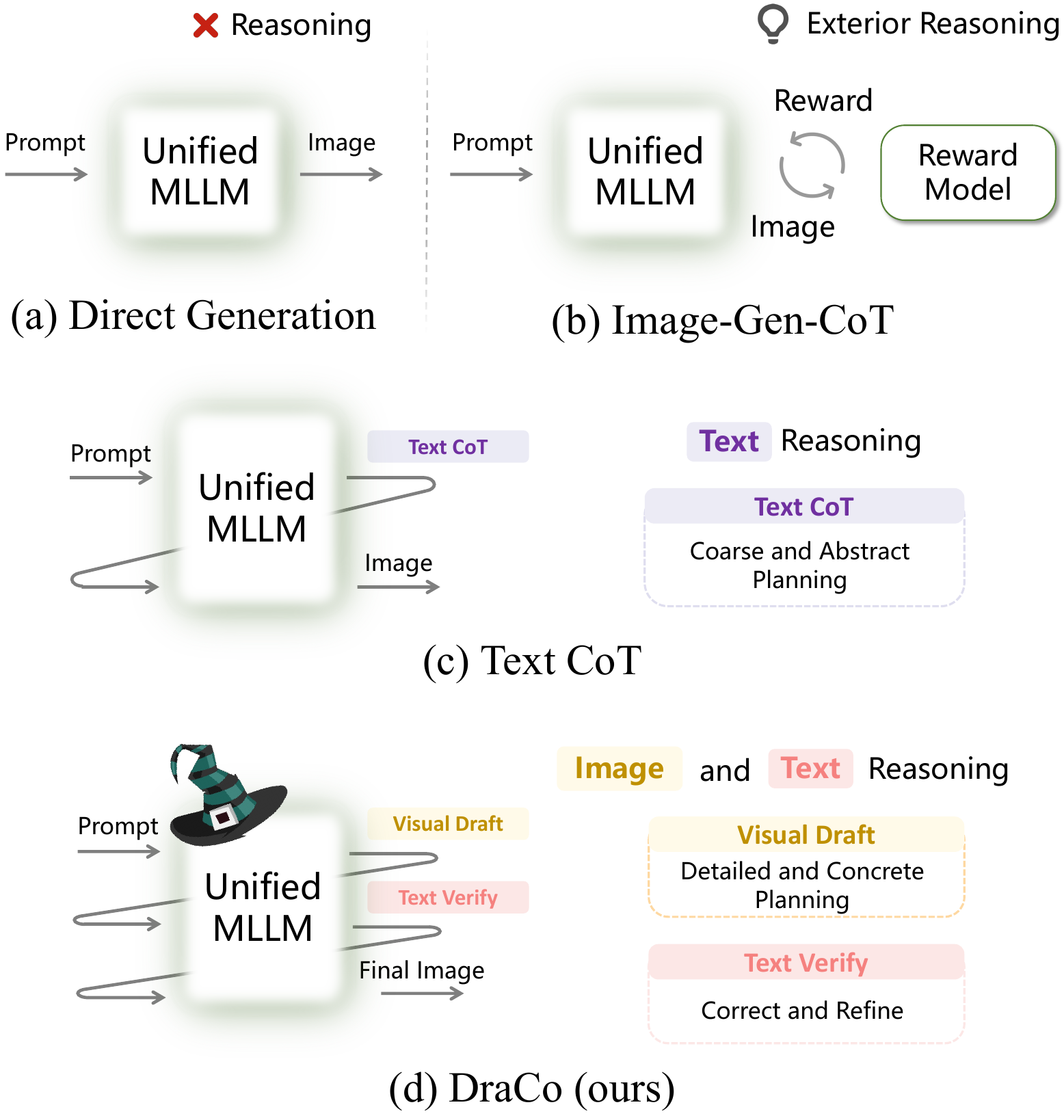}
    \caption{{\bf Conceptual Comparison of CoT Reasoning for T2I Generation.} (a) Generation without reasoning. (b) Employing exterior reward models to guide generation. (c) Generating Text CoT before producing image. (d) {\bf \draco{}}: Producing visual draft for detailed planning and verify it with text reasoning, then correct and refine the draft for final output.}
    \label{fig:intro}
\end{figure}

\begin{figure*}[!t]
    \centering
    \includegraphics[width=\linewidth]{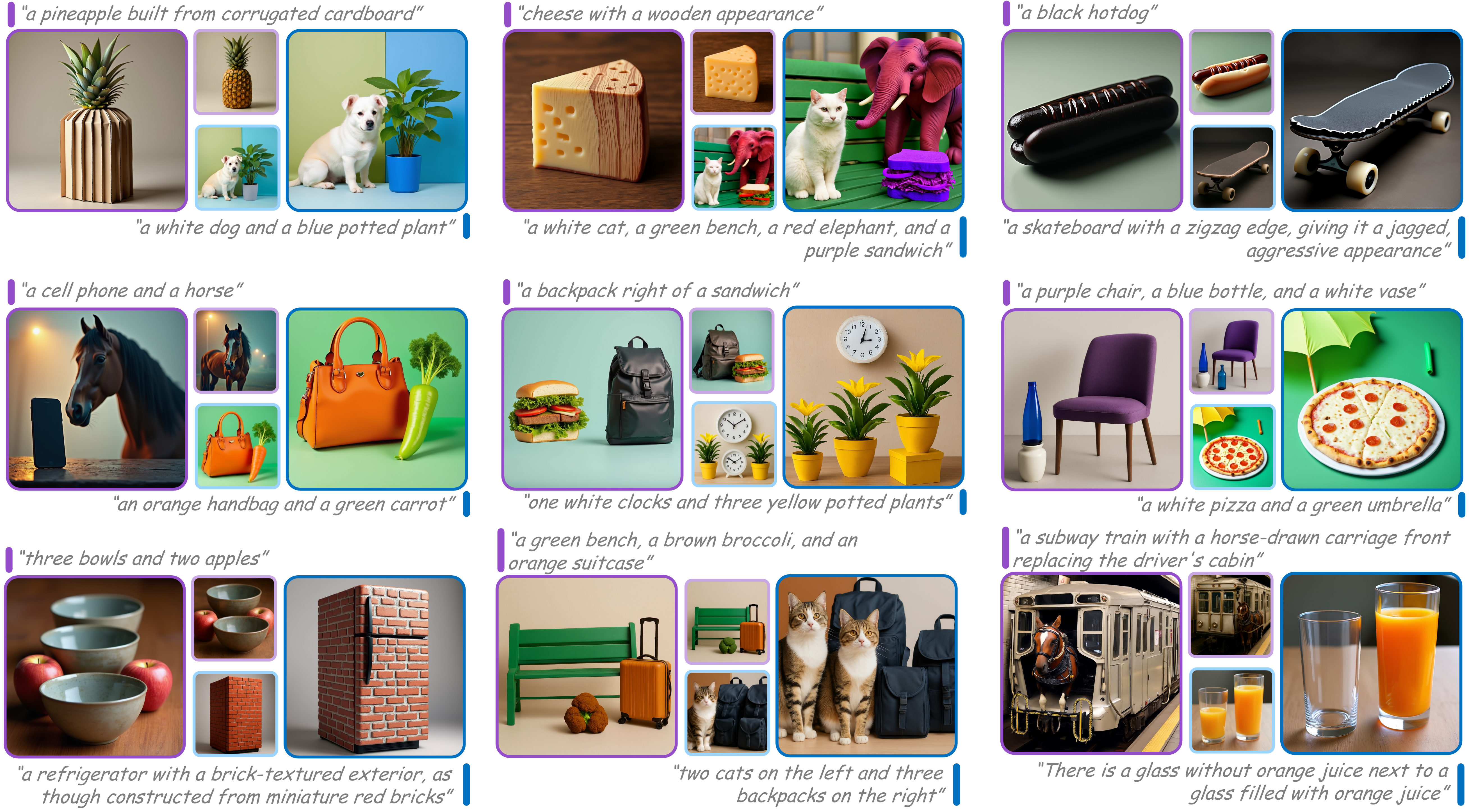}
    \caption{{\bf Visualization of \draco{} Output.} For each example, the larger image represents the final output, while the smaller image is the visual draft. The corresponding prompt is located in the corner of each set.}
    \label{fig:vis2}
\end{figure*} 

Recently, unified multimodal large language models (MLLMs)~\cite{bagel, wu2024janus,wu2024vila,zhou2024transfusion,chen2025blip3o,emu2,xiao2025omnigen,team2024chameleon,liquid,qu2024tokenflow, lin2025uniworld} have emerged as powerful architectures that integrate both visual understanding and generation capabilities, as exemplified by recent works such as Bagel~\cite{bagel}, EMU3~\cite{wang2024emu3}, and Janus~\cite{wu2024janus}. By consolidating these two abilities within a single framework, unified MLLMs have demonstrated remarkable performance and exhibit emergent properties that models possessing only one ability fail to acquire. For instance, unified MLLMs can understand interleaved context including both images and texts and subsequently produce images according to the input instruction~\cite{wu2025omnigen2,xin2025lumina}.

Concurrently, advances in chain-of-thought (CoT) reasoning~\cite{wei2022chain, kojima2022large} have shown remarkable success across various domains, including mathematical problems~\cite{amini2019mathqa,hendrycksmath2021, Lu2023MathVistaEM, zhang2024mathverse, zhang2024mavis}, visual reasoning~\cite{yue2023mmmu, jiang2025mme, chen2025r1v, zhan2025visionr1, meng2025mm}, and multi-agent systems~\cite{mai2025agent, wei2025webagent}. While several pioneering works have extended CoT to text-to-image (T2I) generation tasks~\cite{guo2025can, jiang2025cot, bagel, zhang2025reasongen, pan2025focusdiff, gu2025improving} on unified MLLMs, their exploration has not fully exploited the unified architecture of MLLMs. For example, Image-Gen-CoT~\cite{guo2025can} leverages a reward model to evaluate the quality potential of an image during the early generation stage. This paradigm treats unified MLLM merely as a text-to-image generator, only employing its image generation capabilities.
Later, several methods~\cite{bagel, jiang2025cot} have been proposed to generate textual reasoning for the given prompt prior to image synthesis. However, for generating a dense modality like an image, only planning with text offers too vague and coarse guidance. Besides, only textual understanding ability is leveraged for CoT.
This raises a natural question: \textbf{\textit{Can we design a CoT mechanism with both textual and visual content as better planning and verifier for improved text-to-image generation?}}

\begin{figure*}[!t]
    \centering
    \includegraphics[width=\linewidth]{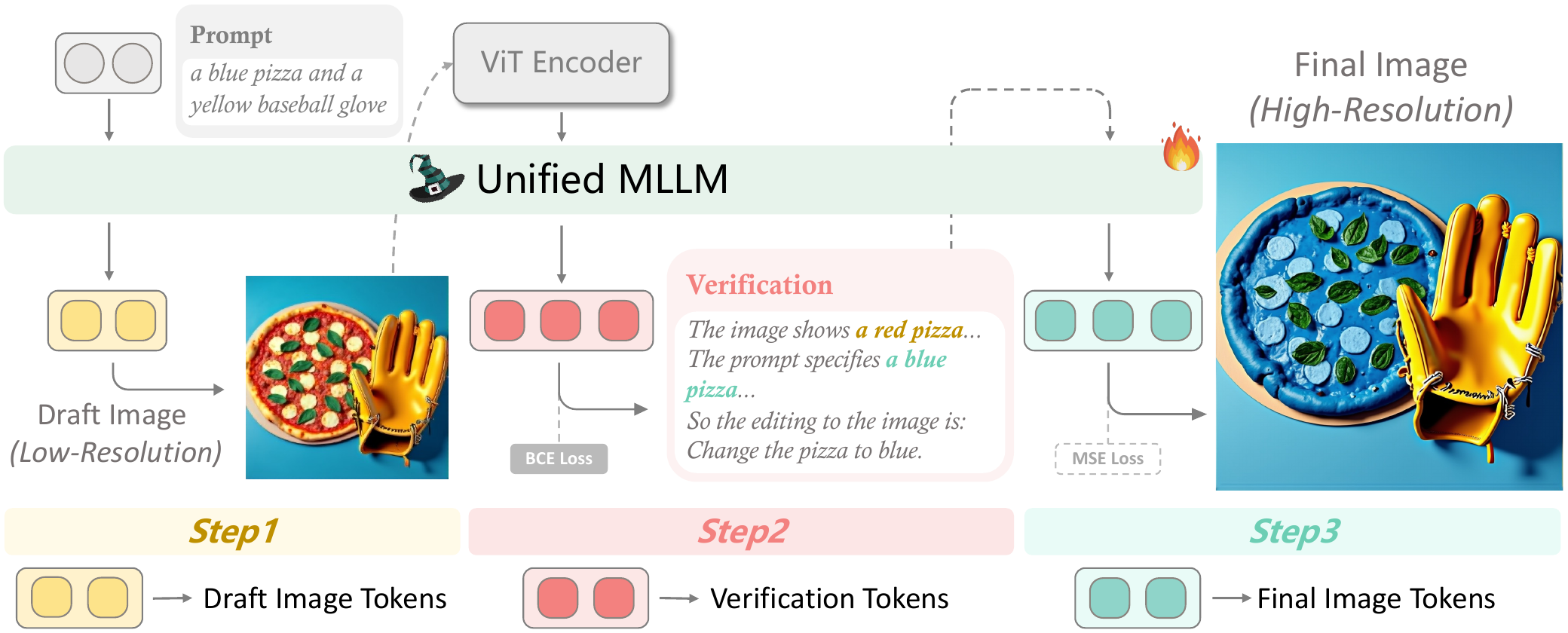}
    \caption{{\bf Framework of \draco{}.} \draco{} contains three steps for generation: draft sketching, draft verification, and corrective refinement.}
    \label{fig:main}
\end{figure*}

To answer this question, we propose \textbf{Draft-as-CoT (\draco{})}, an interleaved CoT reasoning approach that fully leverages the unified framework for the T2I task. 
Our method first generates a low-resolution draft image as visual planning, then leverages the model's inherent understanding capability to verify the draft with any semantic misalignments, and ultimately refines and enriches the draft through selective correction and super-resolution. We show the comparison of \draco{} with previous methods in Fig.~\ref{fig:intro}.

Our method is motivated by two key limitations of textual reasoning for T2I generation. First, text is too abstract to design every aspect of an image in detail, especially on low-level features like appearance and styles. In contrast, planning with a draft image could sufficiently display all the essential visual information during planning. Second, reasoning with a draft image provides an opportunity to preview the image to generate.  
This addresses a fundamental challenge where current T2I models suffer: difficulty in producing the correct image in one pass. Specifically, due to the natural distribution of real-world data, rare attributes or object combinations are significantly underrepresented in training datasets~\cite{samuel2024generating, chen2025failureatlas}. For example, when prompted with ``a white orange'', models often struggle because they have learned to strongly associate the object ``orange'' with its typical color attribute rather than treating them as independent concepts. This binding leads to systematic failures in generating unusual yet valid attribute combinations. However, by previewing the draft, we do not force the model to directly generate the perfect image. Instead, we make the unified MLLM itself identify and refine its flawed planning for the final output.

To facilitate our proposed \draco{} reasoning paradigm, the unified MLLM must be capable of identifying mistakes and precisely controlling and operating on drafts to ensure successful corrections. Since no existing dataset provides such capabilities, we carefully curate a training dataset, \draco{}-240K, targeting three atomic correction capabilities. For each capability, we design a rigorous data synthesizing pipelines that full exploits the synergy between the MLLM~\cite{qwen3technicalreport}, advanced editing models~\cite{labs2025flux1kontextflowmatching}, and segmentation models~\cite{ren2024grounded}.
Additionally, how to conduct classifier-free guidance (CFG)~\cite{ho2022classifier} for interleaved reasoning remains an open question. We propose \draco{}-CFG, a specific type of CFG for \draco{} to explicitly strengthen the two major conditions for final generation: visual semantics from the draft image and correction instructions from verification. 

Experimental results show that \draco{} significantly enhances a strong baseline, Bagel, achieving an 8\% improvement on GenEval~\cite{ghosh2023geneval} and surpassing the text-CoT-based method by 4\%. We find that \draco{} consistently demonstrates superior performance on more challenging benchmarks such as ImagineBench and GenEval++~\cite{ye2025echo}. The visualization output of \draco{} is shown in Fig.~\ref{fig:vis2}.

In summary, our contributions are as follows:
\begin{itemize}
    \item \textbf{A novel interleaved reasoning paradigm:} We introduce Draft-as-CoT, a new interleaved reasoning mechanism that incorporates detailed visual planning and preview for rare concept and combination generation.

    \item \textbf{A comprehensive training dataset with automated pipeline:} We design and construct \draco{}-240K to empower the model with more precise control and diverse operations to correct the drafts following verifications.
    
    \item \textbf{A CFG method for interleaved reasoning:} We present a CFG strategy designed specifically for \draco{}, enabling explicit emphasis on different conditioning signals.
    
\end{itemize}

\section{Method}

We first introduce the structure of our adopted unified MLLM, Bagel, in Section~\ref{method: preliminary}. Then we elaborate on the design idea and pipeline of \draco{} in Section~\ref{method: Draft-as-CoT}. Finally, we introduce the construction of \draco{}-240K in Section~\ref{method: training_dataset}.

\subsection{Preliminary}
\label{method: preliminary}
Our work is built upon a popular unified MLLM, Bagel, which preserves both visual generation capability, including text-to-image generation and image editing, and visual understanding capability. Bagel is composed of three main components: a ViT vision encoder~\cite{tschannen2025siglip} for understanding, a VAE encoder~\cite{kingma2013auto} for generation, and two transformers~\cite{vaswani2017attention} forming a Mixture-of-Transformer-Experts (MoT). One branch of transformer specifically handles VAE tokens, while another branch processes ViT tokens and text tokens. 
For visual understanding tasks, an image is first encoded by ViT into a sequence of image tokens, which are then fed into the transformer to autoregressively generate the text tokens.
For visual generation tasks, Bagel adopts the Rectified Flow~\cite{lipman2022flow, liu2022flow, esser2024scaling} method to produce VAE tokens.

\subsection{Draft-as-CoT}
\label{method: Draft-as-CoT}
We present the pipeline of \draco{} in Fig.~\ref{fig:main}. The entire process can be split into three steps: draft sketching, draft verification, and corrective refinement.

\paragraph{Draft Sketching.}
The goal of the first step in \draco{} is to form a sketch of the specified semantics in the prompt, including foregrounds like objects with their attributes and relations, and holistic information like backgrounds and image styles.
Bagel is typically employed to generate high-resolution images, such as 1024$\times$1024. While these images exhibit excellent details, such granularity is unnecessary for the visual planning stage. Since models often struggle to directly generate images that align with the prompt, subsequent modifications of the initially generated images are still required. Therefore, we only need the model to first present the basic semantics of the intended final image, with other visual details to be supplemented in a later process. Consequently, we propose using Bagel to generate lower-resolution images in the first step, such as 384$\times$384. This resolution is sufficiently small to ensure efficient draft generation compared to generating 1024$\times$1024 images, while being large enough to explicitly display all semantic information, particularly smaller objects.
Thus, in the first step of \draco{}, we only input prompt $p$ to generate a low-resolution draft image.

\paragraph{Draft Verification.}
Leveraging the exceptional visual understanding capabilities inherent in unified MLLMs, we encode the generated draft image through the ViT encoder and re-input it into the unified MLLM. We then instruct the unified MLLM to generate draft verification $v$: first comprehensively understanding the image content, then comparing it with the prompt, and if misalignments are detected, finally summarizing what edits should be applied to the image.
Notably, in the original design of Bagel for understanding interleaved context, such as in editing tasks, it simultaneously inputs both ViT and VAE features. The ViT features facilitate understanding of high-level information, while VAE features maintain low-level details. In contrast, we propose inputting only the ViT features of the draft image without the VAE features.
This design choice stems from the fundamental difference between our task and editing tasks. Editing tasks operate at the same resolution, modifying corresponding elements according to editing instructions while preserving all other information unchanged. Under this requirement, the model requires low-level information to maintain image consistency.
Conversely, our design focuses solely on the high-level semantics of the draft image. 
We could ignore details of irrelevant elements and even irrelevant aspects of prompt-related objects. Dropping the VAE feature could eliminate the constraints from the low-level feature and facilitate more substantial changes.

\begin{figure*}[tp]
    \centering
    \includegraphics[width=\linewidth]{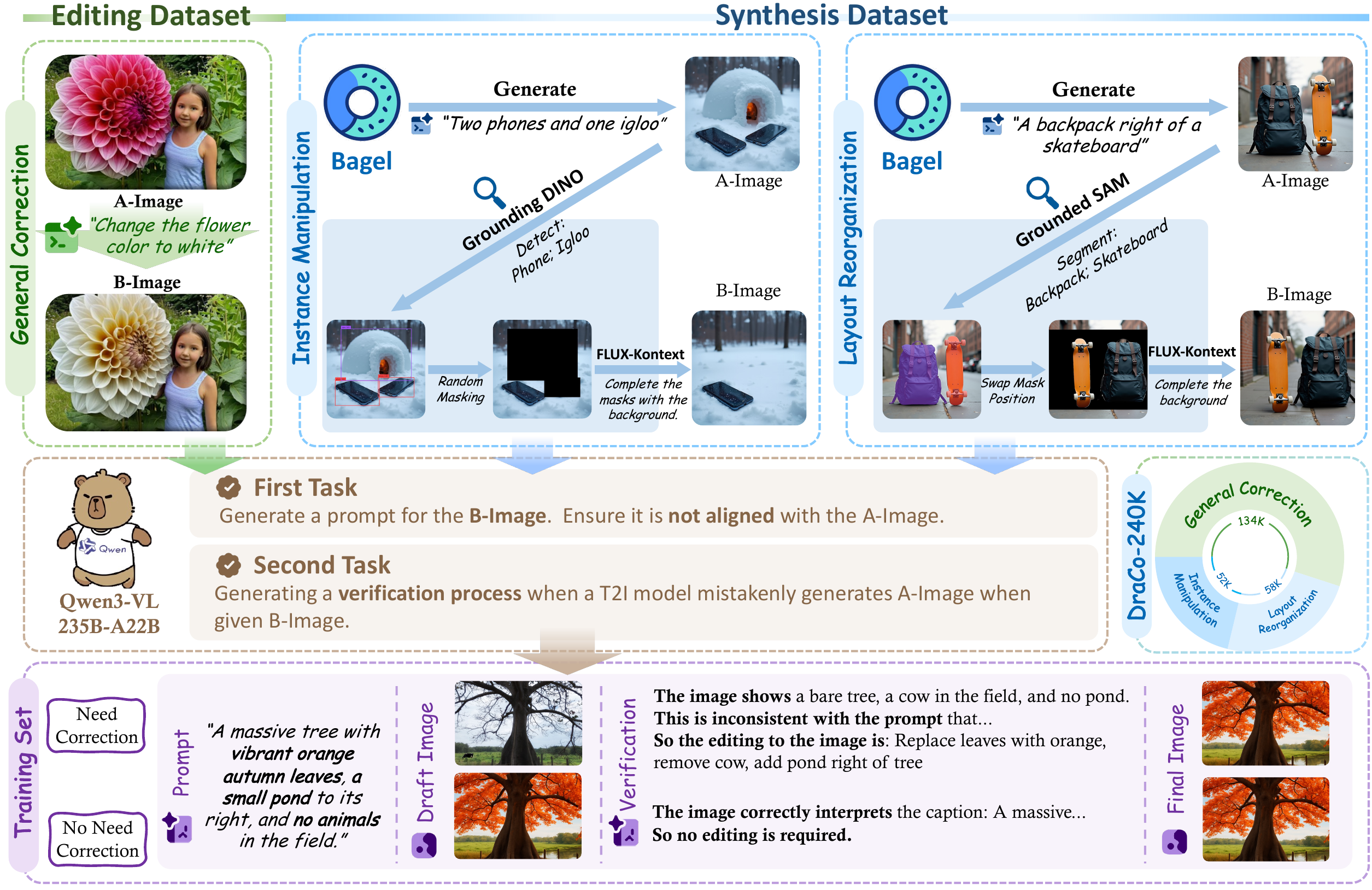}
    \caption{{\bf Construction Pipeline and Examples of \draco{}-240K.} We design specialized data pipelines for each of the three atomic correction capabilities: general correction, instance manipulation, and layout reorganization. We then employ Qwen3-VL~\cite{qwen3technicalreport} to generate prompts and verifications based on the collected image pairs. Finally, we organize the data into two categories for training: corrections needed and corrections not needed.}
    \label{fig:dataset}
\end{figure*}

\paragraph{Corrective Refinement with \draco{}-CFG.}

Eventually, the unified MLLM is required to generate the final high-resolution, detail-rich image aligned with the prompt. This involves both upscaling the draft with improved details and correcting any mistakes detected in the verification.
However, since multiple conditions exist during final image generation (prompt, draft image, and verification), a notable question arises regarding how to organize classifier-free guidance (CFG) to adequately emphasize different conditions in the multimodal context. We identify two core conditions that should control final image generation: one from the draft image's visual condition, constraining the final image to maintain relevant semantic consistency with the draft, and another from the prompt and self-reflection, constraining the image to be modified according to the verification and generated following the prompt. Therefore, with unified model $m$ and noise $x_t$, we design three forms of multimodal context for CFG: the unconditional input $m(\phi,\phi,\phi)$, the draft-only input $m(\phi,{vit},\phi)$, and the fully-conditioned input $m(p, {vit}, v)$. Notably, following our CFG design, we specifically incorporate $m(\phi,{vit},\phi)$ during training, where we train the model to output a high-resolution version of the draft image.
Concretely, we employ the following CFG formulation during inference:

\begin{equation}
\begin{split}
    \hat{m}(p,vit,v)&=m(\phi,\phi,\phi) \\ 
    &+ s_{\text{draft}} \cdot \underbrace{(m(\phi,{vit},\phi)-m(\phi,\phi,\phi))}_{\text{draft condition}} \\ 
    &+ s_{\text{text}} \cdot \underbrace{(m(p, {vit}, v)-m(\phi,{vit},\phi))}_{\text{prompt and correction condition}}.
\end{split}
\end{equation}

\subsection{{\bf \draco{}}-240K}
\label{method: training_dataset}

The correction capability is the most critical among all atom capabilities in \draco{}, integrating super-resolution and advanced editing to generate final images. We first introduce the necessity of constructing the dataset in Section~\ref{dataset: necessity}. Then we detail the two stages with generation pipeline for each capability in Section~\ref{dataset: details}. Finally, we illustrate how to organize the collected data for training in Section~\ref{dataset: training}.

\subsubsection{Necessity of Dataset Construction}
\label{dataset: necessity}

Our zero-shot pilot study reveals that Bagel struggles with correction in two key ways: (1) Despite receiving correct verification, the model fails to follow correction instructions. Despite its inherent editing capability, the model still lacks the precise object control and unconventional editing approaches needed for correcting flawed drafts, like image layout adjustment. (2) Rather than adhering to draft semantics, Bagel generates entirely new images. This undermines our core motivation of using drafts to reduce one-shot generation difficulty.

Existing datasets cannot address these challenges. Therefore, we curate a training dataset for \draco{} with over 240K interleaved reasoning instances. We identify three essential atomic capabilities required for correction and then design data curation pipeline for each one. Throughout data construction, we ensure semantic correlation between draft and final images, ensuring the model to learn to modify only the erroneous elements and retain correct ones.

\begin{figure*}[tp]
    \centering
    \includegraphics[width=\linewidth]{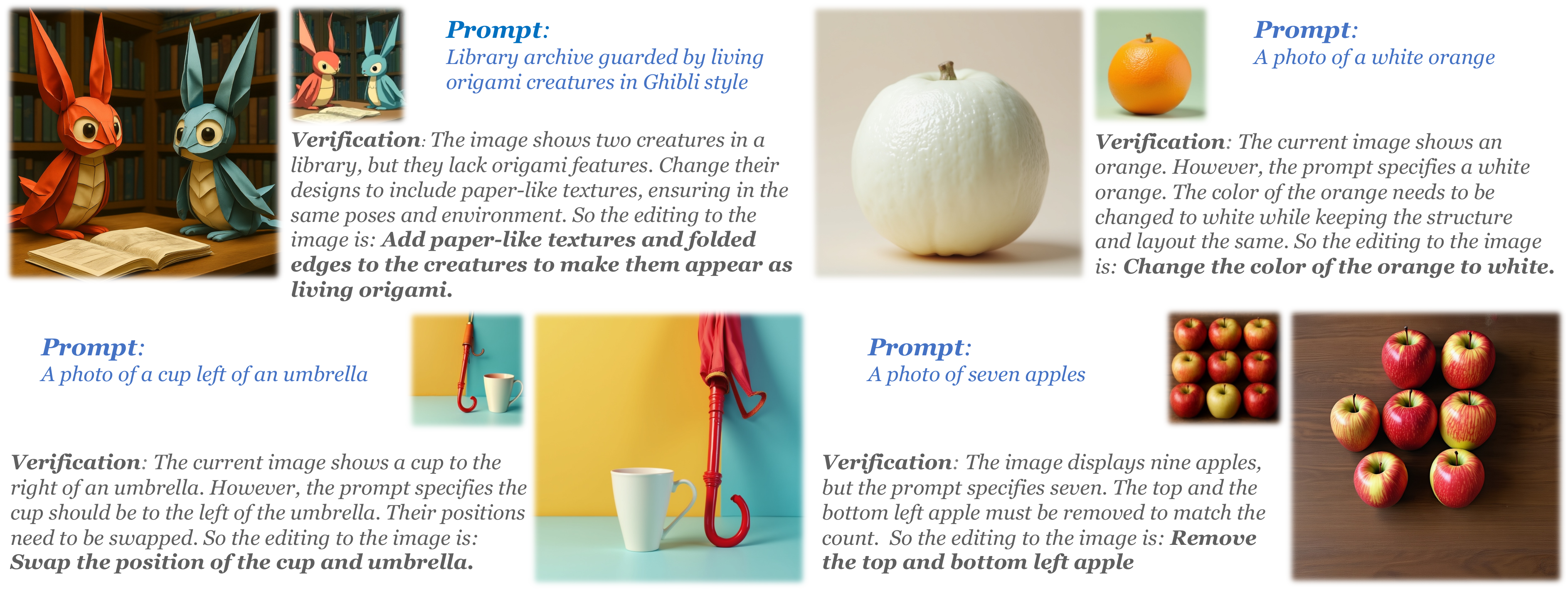}
    \caption{{\bf Detailed Visualization of \draco{} Output.} We showcase the prompt, verification, draft (smaller image), and final output (larger image). \draco{} successfully identifies the misalignment within the draft and conducts the correction based on the suggested modification.}
    \label{fig:vis1}
\end{figure*}

\subsubsection{Construction Details}
\label{dataset: details}

As illustrated in Fig.~\ref{fig:dataset}, the dataset construction consists of two stages: The first stage collects image pairs where the two images differ in some semantic aspects but maintain overall consistency. These images serve as the draft and final images during training. The second stage generates the textual prompt and verification based on the image pairs.
We detail the three pipelines in the first stage below:

\textbullet\ \textbf{General Correction.}
Existing editing datasets are natural sources for general correction data. We denote the pre-editing image as A-Image and the post-editing image as B-Image. The difference between them exists only in the edited objects, while all other characteristics remain identical, perfectly satisfying our data requirements.

\textbullet\ \textbf{Instance Manipulation.}
Instance Manipulation refers to finer-grained control over objects in the image, which is crucial for correction. For example, if the prompt specifies generating three dogs but the model generates five, the verification will likely require removing the two dogs on the left. However, we observe that Bagel struggles to edit instances of the same category, often leading to misinterpretations or incorrect modifications. To address this, we design an automated data pipeline for similar scenarios. We first synthesize numerous prompts specifying multiple instances of the same object, then use Bagel to generate multiple images for each prompt. Next, we extract the objects in the prompt to GroundingDINO~\cite{liu2023grounding} to obtain their bounding boxes. Then, we randomly select some of them and mask them with black rectangles. We input this masked image to FLUX-Kontext~\cite{labs2025flux1kontextflowmatching}, instructing it to inpaint all black masks using the background, thereby obtaining the B-Image. This yields image pairs with multiple identical objects but different quantities. Additionally, we leverage the RefEdit dataset~\cite{pathiraja2025refedit}. Due to its relatively poor quality, we use GPT-4o~\cite{openai2024gpt4o} to regenerate the edited images. These image pairs feature multiple identical objects, but one object differs in certain attributes across the images.

\textbullet\ \textbf{Layout Reorganization.}
As for image layout understanding, we find that Bagel lacks the ability to comprehend spatial relationships in editing instructions, let alone modify them for correction. Therefore, we specifically design a subset for spatial correction. Similarly, we use prompts specifying spatial layouts to have Bagel generate numerous images. We then feed the objects from the prompt into GroundedSAM~\cite{ren2024grounded} to obtain segmentation masks for each object. We randomly swap the positions of these segmentation masks and paste them onto a pure black background. Subsequently, if there is no overlap with the original masks, we paste the periphery of the original image onto the black background to make the newly generated image's background as consistent as possible with the original image. Finally, we use FLUX-Kontext to inpaint the black regions based on the background around the image's periphery, obtaining B-Image, where the number and appearance of each object remains same but the layout is changed.

In the second stage, we randomly swap A-Image and B-Image in the image pairs and input them to Qwen3-VL-235B-A22B~\cite{qwen3technicalreport}. The first task is to output a caption of B-Image as the prompt for training, ensuring it is misaligned with A-Image. The second task is to generate the verification, namely, if a text-to-image model generates A-Image given the prompt from task one, how should it be modified to become B-Image. The verification needs to first observe the misalignments between the current image and prompt, and if any exist, provides the modification approach.

\subsubsection{Training Set}
\label{dataset: training}
Finally, we organize the image pairs, the prompt, and the verification outputs from Qwen3-VL into two types of data: modification needed (A-Image as draft) and no modification needed (B-Image as draft). 
During training, we utilize these two types of data, as well as text-to-image data. The text-to-image data is trained on two resolutions: low resolution, same as the draft, and high resolution, same as the final output. This is to ensure the model produces meaningful draft images and preserves its original generation capability. More details and examples of each capability are shown in the Appendix~\ref{sup:dataset}.

\begin{table*}[t]
\centering
\renewcommand{\arraystretch}{1.2}
\scriptsize
\caption{\textbf{Evaluation results on GenEval.} The best results are in \textbf{bold fonts} with the second best \underline{underlined}.}
\resizebox{0.95\linewidth}{!}{
\begin{tabular}{lcccccccc}
\toprule
\textbf{Method} & \textbf{Single Obj.$\uparrow$} & \textbf{Two Obj.$\uparrow$} & \textbf{Counting$\uparrow$} & \textbf{Colors$\uparrow$} & \textbf{Position$\uparrow$} & \textbf{Color Attri.$\uparrow$} & \textbf{Overall$\uparrow$} \\
\midrule
\multicolumn{8}{c}{\textit{Generation Only}} \\
\midrule
LlamaGen~\cite{llamagen2024} & 0.71 & 0.34 & 0.21 & 0.58 & 0.07 & 0.04 & 0.32 \\
SDXL~\cite{podell2023sdxl} & 0.98 & 0.74 & 0.39 & 0.85 & 0.15 & 0.23 & 0.55 \\
DALL-E~3~\cite{dalle3} & 0.96 & 0.87 & 0.47 & 0.83 & 0.43 & 0.45 & 0.67 \\
SD3-Medium~\cite{sd3medium2024} & \underline{0.99} & \underline{0.94} & \underline{0.72} & 0.89 & 0.33 & 0.60 & 0.74 \\
\midrule
\multicolumn{8}{c}{\textit{Unified MLLM}} \\
\midrule
Show-o~\cite{xie2024show} & 0.95 & 0.52 & 0.49 & 0.82 & 0.11 & 0.28 & 0.53 \\
Janus-Pro-7B~\cite{chen2025janus} & \underline{0.99} & 0.89 & 0.59 & \underline{0.90} & \textbf{0.79} & \underline{0.66} & 0.80 \\
BLIP3-o 8B~\cite{chen2025blip3o} & - & - & - & - & - & - & \underline{0.84} \\
BAGEL~\cite{bagel} & \underline{0.99} & \underline{0.94} & \textbf{0.81} & 0.88 & 0.64 & 0.63 & 0.78 \\
\midrule
\multicolumn{8}{c}{\textit{Unified MLLM w/ CoT}} \\
\midrule
Show-o+PARM~\cite{guo2025can} & 0.98 & 0.55 & 0.54 & 0.83 & 0.13 & 0.29 & 0.55 \\
T2I-R1~\cite{jiang2025cot} & \underline{0.99} & 0.91 & 0.53 & {\bf 0.91} & \underline{0.76} & 0.65 & 0.79  \\
BAGEL-Think~\cite{bagel} & \underline{0.99} & \underline{0.94} & \textbf{0.81} & 0.88 & 0.64 & 0.63 & 0.82 \\
\midrule
\textbf{\draco{} (Ours)} & \textbf{1.00} & \textbf{0.99} & \textbf{0.81} & \textbf{0.91} & {0.70} & \textbf{0.76} & \textbf{0.86} \\
\bottomrule
\end{tabular}
}
\label{tab:geneval}
\end{table*}

\subsection{Training Loss}
\label{method: loss}
We conduct supervised fine-tuning from Bagel. For our curated training dataset, we sequentially input the prompt tokens, the ViT feature of draft images, the verifications, and finally the noisy VAE tokens of the final image.
We only calculate Binary Cross Entropy (BCE) loss on the verifications and Mean Squared Error (MSE) on the VAE tokens:

\begin{align}
\mathcal{L}_{\text{verification}} &= -\frac{1}{|v|} \sum_{i=1}^{|v|}  \log(v_i), \\
\mathcal{L}_{\text{final image}} &= \mathbb{E}_{t, {x}_0, {x}_1} \left[ \left\| m(t, {x}_t) - ({x}_1 - {x}_0) \right\|^2 \right].
\end{align}

To support \draco{}-CFG, we apply two dropout strategies during training, each with a 5\% probability: (1) dropping all conditions for unconditional generation, or (2) preserving only the ViT feature, as mentioned in Section~\ref{method: Draft-as-CoT}.

\section{Experiment}

\subsection{Experimental Setting}
\paragraph{Evaluation.}
We evaluate our proposed method on GenEval~\cite{ghosh2023geneval}, GenEval++, and ImagineBench~\cite{ye2025echo}. GenEval consists of six subcategories focusing on object existence, attributes, counting, and positions. Building upon this benchmark, GenEval++ presents more challenging scenarios with complex combinations and stricter scoring criteria. Additionally, we test on ImagineBench, which specifically targets unusual object-attribute combinations where text-to-image models typically struggle. We generate one image per prompt for evaluation, due to the slow inference speed of Bagel and follow the official scoring method in each benchmark. During our evaluation, $s_{draft}$ and $s_{text}$ are set to 2 to 6, respectively.

\paragraph{Training Details.}
We empirically observe that Bagel struggles to generate valid 384$\times$384 images. Therefore, we first conduct a text-to-image fine-tuning stage to empower the model with low-resolution image generation capability before training for \draco{}. Additional details are provided in the Appendix~\ref{sup:exp}.
Subsequently, we perform full-parameter fine-tuning for 16K steps and retain the EMA weights. The learning rate is set to 2e-5 with 2K warmup steps. We freeze the ViT encoder and its connector during training to preserve valid visual encoding of high-level information. For each training iteration, the maximum length of concatenated training samples is set to 36K per GPU. We utilize 8 H800 GPUs for training.

\subsection{Main Results}
We present quantitative results in Tables~\ref{tab:geneval} and \ref{tab:imagine_geneval}, with qualitative results shown in Figures~\ref{fig:vis2} and \ref{fig:vis1}. We compare our method against generation-only models (including diffusion-based and autoregressive models), unified MLLMs, and unified MLLMs with CoT planning.

Our method substantially outperforms other methods with CoT across all three benchmarks. On GenEval, \draco{} achieves an overall score of 0.86 and attains the highest scores in five out of six subtasks. Notably, \draco{} excels particularly in the color attribute subtask, which involves objects with different specified colors. This demonstrates that \draco{} is especially effective at handling complex attribute combinations. As shown in the table, the exterior reward model PARM~\cite{guo2025can} brings 2\% improvement on Show-o~\cite{xie2024show}, from 0.53 to 0.55. However, our method showcases 6\% greater improvement based on a more strong baseline starting from 0.78. On ImagineBench, compared to the baseline Bagel without reasoning, \draco{} achieves a notable improvement of 0.91 points. \draco{} also surpasses text-only reasoning by 0.18 points, indicating that visual drafting and preview are more effective for generating rare attribute combinations. Furthermore, \draco{} also demonstrates advantages on the more challenging GenEval++ with overall score of 0.40. Interestingly, Bagel-Think underperforms vanilla Bagel, suggesting potential limitations of text-based planning. In contrast, \draco{} consistently delivers superior performance across all benchmarks.

\begin{table*}[t]
    \centering
    \caption{\textbf{Evaluation results on Imagine-Bench and GenEval++.} The best results are in \textbf{bold fonts} with the second best \underline{underlined}.}
    \label{tab:imagine_geneval}
    \resizebox{1.0\linewidth}{!}{
        \begin{tabular}{l|cccc|c|ccccccc|c}
            \toprule
                    & \multicolumn{5}{c|}{\textbf{Imagine-Bench}}  &  \multicolumn{8}{c}{\textbf{GenEval++}}   \\
            Method  & Attr. shift$\uparrow$ & Spatiotem.$\uparrow$ & Hybrid.$\uparrow$ & Multi-Obj.$\uparrow$ & {\bf Overall}$\uparrow$ & Color$\uparrow$ & Count$\uparrow$ & Color/Count$\uparrow$ & Color/Pos$\uparrow$ & Pos/Count$\uparrow$ & Pos/Size$\uparrow$ & Multi-Count$\uparrow$ & {\bf Overall}$\uparrow$ \\
            \midrule
            \multicolumn{14}{c}{\textit{Generation Only}} \\
            \midrule
            SDv2.1~\cite{rombach2022high} & 4.46 & 5.06 & 4.12 & 3.49 & 4.30 & 0.00 & 0.33 & 0.03 & 0.00 & 0.00 & 0.03 & 0.08 & 0.06 \\
            SDXL~\cite{podell2023sdxl} & 4.42 & 6.32 & 4.93 & 4.50 & 4.97 & 0.05 & 0.38 & 0.00 & 0.00 & 0.00 & 0.00 & 0.00 & 0.06 \\
            FLUX.1-dev~\cite{labs2025flux1kontextflowmatching} & 5.68 & 7.13 & 6.38 & 5.24 & 6.06 & 0.35 & \textbf{0.63} & 0.15 & 0.28 & 0.20 & 0.38 & 0.23 & 0.31 \\
            \midrule
            \multicolumn{14}{c}{\textit{Unified MLLM}} \\
            \midrule
            Janus-Pro-7B~\cite{chen2025janus} & 5.30 & 7.28 & 6.73 & 6.04 & 6.22 & \underline{0.45} & 0.30 & 0.13 & 0.30 & 0.08 & 0.35 & 0.13 & 0.25 \\
            BLIP3-o 4B~\cite{chen2025blip3o} & 5.48 & 6.79 & 6.93 & 6.09 & 6.23 & 0.13 & 0.23 & 0.10 & \underline{0.45} & 0.13 & \underline{0.55} & 0.23 & 0.26 \\
            BLIP3-o 8B~\cite{chen2025blip3o} & 5.80 & 7.08 & 7.06 & 6.44 & 6.51 & 0.25 & 0.25 & 0.13 & {\bf 0.60} & 0.13 & {\bf 0.58} & 0.23 & 0.31 \\
            Bagel~\cite{bagel} & 5.37 & 6.93 & 6.50 & 6.41 & 6.20 & 0.33 & \underline{0.60} & 0.25 & 0.33 & \underline{0.25} & 0.48 & \textbf{0.38} & \underline{0.37} \\
            \midrule
            \multicolumn{14}{c}{\textit{Unified MLLM w/ CoT}} \\
            \midrule
            T2I-R1~\cite{jiang2025cot} & 5.85 & {\bf 7.70} & 7.36 & 6.68 & 6.78 & {\bf 0.68} & 0.33 & 0.20 & 0.35 & 0.08 & 0.25 & 0.30 & 0.31 \\
            Bagel-Think~\cite{bagel} & \underline{6.26} & 7.13 & \underline{7.74} & \underline{6.96} & \underline{6.93} & 0.35 & 0.38 & {\bf 0.35} & \underline{0.45} & 0.13 & 0.48 & \underline{0.35} & 0.35 \\
            \midrule
            \textbf{\draco{} (Ours)} & {\bf 6.40} & \underline{7.30} & {\bf 7.99} & {\bf 7.20} & {\bf 7.11} & \underline{0.45} & 0.53 & \underline{0.28} & 0.40 & {\bf 0.28} & {0.53} & \textbf{0.38} & {\bf 0.40} \\
            \bottomrule
        \end{tabular}
    }
\end{table*}

\subsection{Ablation Study}
\begin{table}[!t]
\caption{\textbf{Ablation Results of Key Designs in \draco{}.}}
\centering
\begin{tabular}{cccc}
\toprule
Draft & \multirow{2}{*}{CFG} & Draft & {GenEval} \\
Resolution & & VAE Input & Overall $\uparrow$ \\
\midrule
128 & \draco{}-CFG & \ding{55} & 0.76  \\
1024 & \draco{}-CFG & \ding{55} & 0.75 \\
384 & Original & \ding{55} & 0.83 \\
384 & \draco{}-CFG & $\checkmark$ & 0.84 \\
384 & \draco{}-CFG & \ding{55} & 0.86 \\
\midrule
\end{tabular}
\vspace{-0.3cm}
\label{table:ablation}
\end{table}

\begin{figure}[!t]
  \centering
  \includegraphics[width=\linewidth]{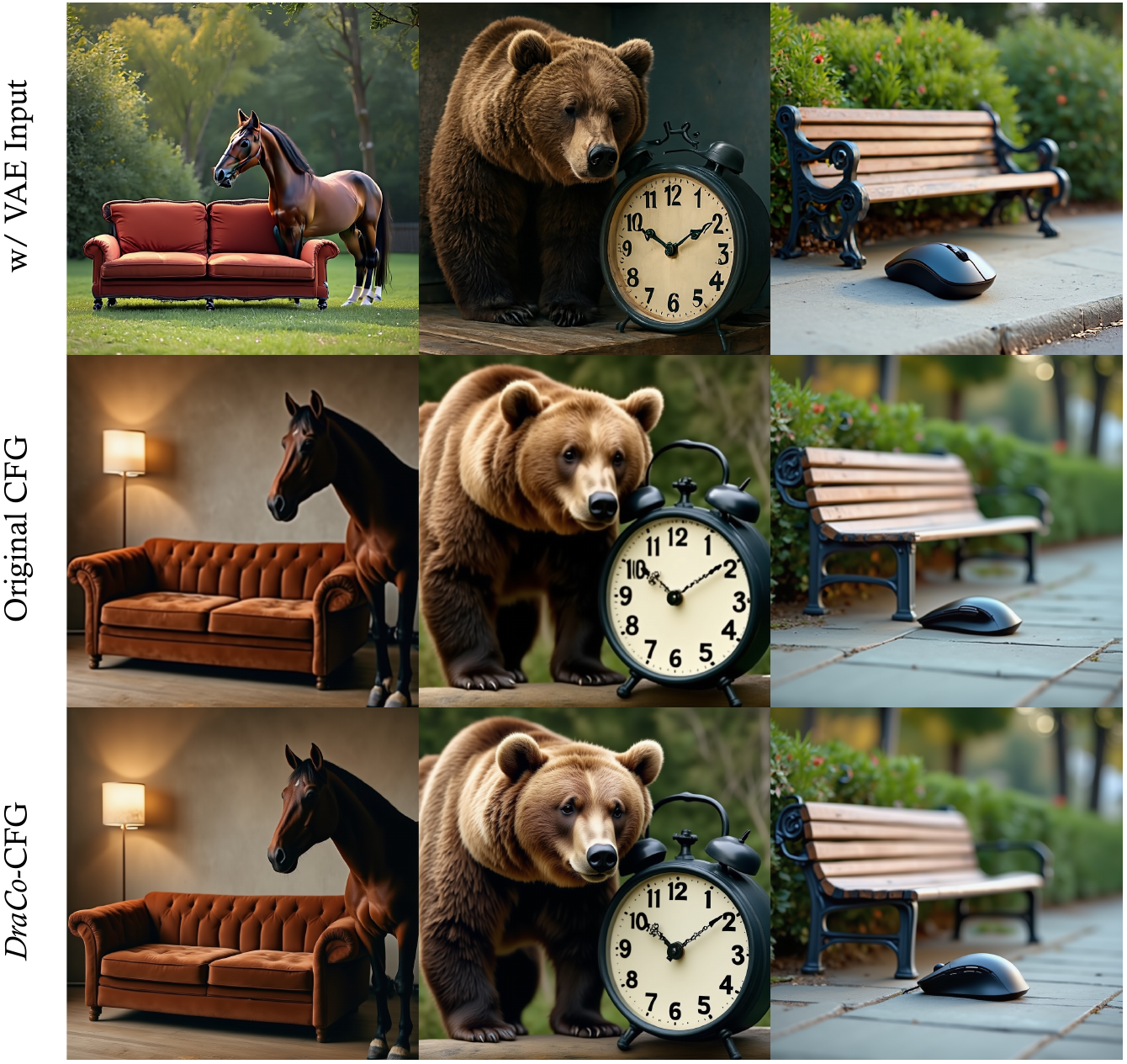}
  \caption{\textbf{Qualitative Comparison Between Imputing VAE Feature of the Draft, Original CFG and \draco{}-CFG.} Zoom in for better visualization.}
  \label{fig:draco_cfg}  
  \vspace{-0.5cm}
\end{figure}

\paragraph{Effectiveness of \draco{}-CFG.}
We demonstrate the effectiveness of our proposed \draco{}-CFG by comparing it with the original CFG method adopted by Bagel, with details provided in the Appendix~\ref{sup:dracocfg}. As shown in Table~\ref{table:ablation} and Fig.~\ref{fig:draco_cfg}, \draco{}-CFG exhibits superior performance both quantitatively and qualitatively. Quantitatively, \draco{}-CFG surpasses the original CFG by 3\% on the overall score of GenEval. Qualitatively, compared to the blurriness of the original CFG, \draco{}-CFG demonstrates significantly improved clarity and high quality, as evidenced by the details such as the clock figures.

\paragraph{Is the VAE feature of the draft needed?}
We investigate the necessity of incorporating VAE features from the draft alongside ViT features during training. Results in Table~\ref{table:ablation} show that including VAE features yields an overall score of 0.84, which is 2\% lower than the model without VAE features. This suggests that low-level features impose constraints on the final correction step. 
Qualitatively, as shown in Fig.~\ref{fig:draco_cfg}, we observe that VAE features may introduce artifacts in the final output due to over-adherence to the draft, including unnatural lighting on the horse, improper intersection between the horse's legs and the sofa, and disconnected frame at the top of the clock.
 
\paragraph{Draft Resolution.}
We examine the impact of draft resolution on final results by testing resolutions of 128$\times$128 and 1024$\times$1024, as shown in Table~\ref{table:ablation}. At 128$\times$128 resolution, the model cannot adequately express its planning in the excessively small draft. Besides, the resolution is too small for the model to clearly identify the image for the verification step. These problems result in poor performance with a score of 0.76. Conversely, the excessively large resolution of 1024$\times$1024 substantially increases the token length per training sample, thereby reducing the number of training samples within the same training iterations. This also yields unsatisfactory results, with a score of 0.75.

\section{Conclusion}

In this work, we introduce \textbf{\draco{}}, a novel interleaved reasoning paradigm that fully exploits the unified architecture of MLLMs by incorporating both visual and textual chain-of-thought for text-to-image generation. 
\draco{} addresses the coarse-grained nature of textual planning and enables generation of rare attribute combinations that models typically struggle with due to training data biases.
We also curate \draco{}-240K with specialized automated pipelines targeting three atomic correction capabilities, and propose \draco{}-CFG to explicitly strengthen visual semantics from drafts and correction instructions from verification. 
Extensive experiments demonstrate \draco{}'s superiority, achieving +8\% on GenEval, +0.91 on ImagineBench, and +3\% on GenEval++. 
Our work offers new direction for enhancing visual generation tasks through interleaved multimodal reasoning in unified MLLMs.

{
    \small
    \bibliographystyle{ieeenat_fullname}
    \bibliography{main}

@String(ECCV= {Eur. Conf. Comput. Vis.})

@String(AAAI = {AAAI})

@String(ECCV  = {ECCV})

@article{zhou2024transfusion,
  title={Transfusion: Predict the next token and diffuse images with one multi-modal model},
  author={Zhou, Chunting and Yu, Lili and Babu, Arun and Tirumala, Kushal and Yasunaga, Michihiro and Shamis, Leonid and Kahn, Jacob and Ma, Xuezhe and Zettlemoyer, Luke and Levy, Omer},
  journal={arXiv preprint arXiv:2408.11039},
  year={2024}
}

@article{liquid,
  title={Liquid: Language models are scalable multi-modal generators},
  author={Wu, Junfeng and Jiang, Yi and Ma, Chuofan and Liu, Yuliang and Zhao, Hengshuang and Yuan, Zehuan and Bai, Song and Bai, Xiang},
  journal={arXiv preprint arXiv:2412.04332},
  year={2024}
}

@article{chen2025blip3o,
  title={BLIP3-o: A Family of Fully Open Unified Multimodal Models-Architecture, Training and Dataset},
  author={Chen, Jiuhai and Xue, Le and Shu, Manli and Zhang, Yunhao and Zhang, Yu and Xu, Ran and Wang, Xin Eric and Xiong, Caiming},
  journal={arXiv preprint arXiv:2505.09568},
  year={2025}
}

@article{chen2025janus,
  title={Janus-Pro: Unified Multimodal Understanding and Generation with Data and Model Scaling},
  author={Chen, Xiaokang and Wu, Zhiyu and Liu, Xingchao and Pan, Zizheng and Liu, Wen and Xie, Zhenda and Yu, Xingkai and Ruan, Chong},
  journal={arXiv preprint arXiv:2501.17811},
  year={2025}
}

@article{wang2024emu3,
  title={Emu3: Next-token prediction is all you need},
  author={Wang, Xinlong and Zhang, Xiaosong and Luo, Zhengxiong and Sun, Quan and Cui, Yufeng and Wang, Jinsheng and Zhang, Fan and Wang, Yueze and Li, Zhen and Yu, Qiying and others},
  journal={arXiv preprint arXiv:2409.18869},
  year={2024}
}

@article{qu2024tokenflow,
  title={Tokenflow: Unified image tokenizer for multimodal understanding and generation},
  author={Qu, Liao and Zhang, Huichao and Liu, Yiheng and Wang, Xu and Jiang, Yi and Gao, Yiming and Ye, Hu and Du, Daniel K and Yuan, Zehuan and Wu, Xinglong},
  journal={arXiv preprint arXiv:2412.03069},
  year={2024}
}

@article{xie2024show,
  title={Show-o: One single transformer to unify multimodal understanding and generation},
  author={Xie, Jinheng and Mao, Weijia and Bai, Zechen and Zhang, David Junhao and Wang, Weihao and Lin, Kevin Qinghong and Gu, Yuchao and Chen, Zhijie and Yang, Zhenheng and Shou, Mike Zheng},
  journal={arXiv preprint arXiv:2408.12528},
  year={2024}
}

@article{dong2023dreamllm,
  title={Dreamllm: Synergistic multimodal comprehension and creation},
  author={Dong, Runpei and Han, Chunrui and Peng, Yuang and Qi, Zekun and Ge, Zheng and Yang, Jinrong and Zhao, Liang and Sun, Jianjian and Zhou, Hongyu and Wei, Haoran and others},
  journal={arXiv preprint arXiv:2309.11499},
  year={2023}
}

@article{team2024chameleon,
  title={Chameleon: Mixed-modal early-fusion foundation models},
  author={Team, Chameleon},
  journal={arXiv preprint arXiv:2405.09818},
  year={2024}
}

@article{wei2022chain,
  title={Chain-of-thought prompting elicits reasoning in large language models},
  author={Wei, Jason and Wang, Xuezhi and Schuurmans, Dale and Bosma, Maarten and Xia, Fei and Chi, Ed and Le, Quoc V and Zhou, Denny and others},
  journal={Advances in neural information processing systems},
  volume={35},
  pages={24824--24837},
  year={2022}
}

@article{chen2025mint,
  title={MINT-CoT: Enabling Interleaved Visual Tokens in Mathematical Chain-of-Thought Reasoning},
  author={Chen, Xinyan and Zhang, Renrui and Jiang, Dongzhi and Zhou, Aojun and Yan, Shilin and Lin, Weifeng and Li, Hongsheng},
  journal={arXiv preprint arXiv:2506.05331},
  year={2025}
}

@article{jiang2025mme,
  title={MME-CoT: Benchmarking Chain-of-Thought in Large Multimodal Models for Reasoning Quality, Robustness, and Efficiency},
  author={Jiang, Dongzhi and Zhang, Renrui and Guo, Ziyu and Li, Yanwei and Qi, Yu and Chen, Xinyan and Wang, Liuhui and Jin, Jianhan and Guo, Claire and Yan, Shen and others},
  journal={arXiv preprint arXiv:2502.09621},
  year={2025}
}

@misc{chen2025r1v,
  author       = {Chen, Liang and Li, Lei and Zhao, Haozhe and Song, Yifan and Vinci},
  title        = {R1-V: Reinforcing Super Generalization Ability in Vision-Language Models with Less Than \$3},
  howpublished = {\url{https://github.com/Deep-Agent/R1-V}},
  note         = {Accessed: 2025-02-02},
  year         = {2025}
}

@article{zhan2025visionr1,
  title={Vision-R1: Evolving Human-Free Alignment in Large Vision-Language Models via Vision-Guided Reinforcement Learning},
  author={Zhan, Yufei and Zhu, Yousong and Zheng, Shurong and Zhao, Hongyin and Yang, Fan and Tang, Ming and Wang, Jinqiao},
  journal={arXiv preprint arXiv:2503.18013},
  year={2025}
}

@article{zhang2024mavis,
  title={MAVIS: Mathematical Visual Instruction Tuning},
  author={Zhang, Renrui and Wei, Xinyu and Jiang, Dongzhi and Zhang, Yichi and Guo, Ziyu and Tong, Chengzhuo and Liu, Jiaming and Zhou, Aojun and Wei, Bin and Zhang, Shanghang and others},
  journal={arXiv preprint arXiv:2407.08739},
  year={2024}
}

@article{wei2025webagent,
  title={Webagent-r1: Training web agents via end-to-end multi-turn reinforcement learning},
  author={Wei, Zhepei and Yao, Wenlin and Liu, Yao and Zhang, Weizhi and Lu, Qin and Qiu, Liang and Yu, Changlong and Xu, Puyang and Zhang, Chao and Yin, Bing and others},
  journal={arXiv preprint arXiv:2505.16421},
  year={2025}
}

@article{mai2025agent,
  title={Agent rl scaling law: Agent rl with spontaneous code execution for mathematical problem solving},
  author={Mai, Xinji and Xu, Haotian and Li, Zhong-Zhi and Wang, Weinong and Hu, Jian and Zhang, Yingying and Zhang, Wenqiang and others},
  journal={arXiv preprint arXiv:2505.07773},
  year={2025}
}

@article{zhang2024mathverse,
  title={Mathverse: Does your multi-modal llm truly see the diagrams in visual math problems?},
  author={Zhang, Renrui and Jiang, Dongzhi and Zhang, Yichi and Lin, Haokun and Guo, Ziyu and Qiu, Pengshuo and Zhou, Aojun and Lu, Pan and Chang, Kai-Wei and Gao, Peng and others},
  journal={ECCV 2024},
  year={2024}
}

@article{guo2025can,
  title={Can We Generate Images with CoT? Let's Verify and Reinforce Image Generation Step by Step},
  author={Guo, Ziyu and Zhang, Renrui and Tong, Chengzhuo and Zhao, Zhizheng and Gao, Peng and Li, Hongsheng and Heng, Pheng-Ann},
  journal={arXiv preprint arXiv:2501.13926},
  year={2025}
}

@article{Lu2023MathVistaEM,
  title={MathVista: Evaluating Math Reasoning in Visual Contexts with GPT-4V, Bard, and Other Large Multimodal Models},
  author={Pan Lu and Hritik Bansal and Tony Xia and Jiacheng Liu and Chun-yue Li and Hannaneh Hajishirzi and Hao Cheng and Kai-Wei Chang and Michel Galley and Jianfeng Gao},
  journal={ArXiv},
  year={2023},
  volume={abs/2310.02255},
}

@article{gu2025improving,
  title={Improving Chain-of-Thought Efficiency for Autoregressive Image Generation},
  author={Gu, Zeqi and Georgopoulos, Markos and Dai, Xiaoliang and Ghazvininejad, Marjan and Wang, Chu and Juefei-Xu, Felix and Li, Kunpeng and Shi, Yujun and He, Zecheng and He, Zijian and others},
  journal={arXiv preprint arXiv:2510.05593},
  year={2025}
}

@article{ho2022classifier,
  title={Classifier-free diffusion guidance},
  author={Ho, Jonathan and Salimans, Tim},
  journal={arXiv preprint arXiv:2207.12598},
  year={2022}
}

@article{wu2024janus,
  title={Janus: Decoupling visual encoding for unified multimodal understanding and generation},
  author={Wu, Chengyue and Chen, Xiaokang and Wu, Zhiyu and Ma, Yiyang and Liu, Xingchao and Pan, Zizheng and Liu, Wen and Xie, Zhenda and Yu, Xingkai and Ruan, Chong and others},
  journal={arXiv preprint arXiv:2410.13848},
  year={2024}
}

@article{ghosh2023geneval,
  title={Geneval: An object-focused framework for evaluating text-to-image alignment},
  author={Ghosh, Dhruba and Hajishirzi, Hannaneh and Schmidt, Ludwig},
  journal={Advances in Neural Information Processing Systems},
  volume={36},
  pages={52132--52152},
  year={2023}
}

@inproceedings{rombach2022high,
  title={High-resolution image synthesis with latent diffusion models},
  author={Rombach, Robin and Blattmann, Andreas and Lorenz, Dominik and Esser, Patrick and Ommer, Bj{\"o}rn},
  booktitle={Proceedings of the IEEE/CVF conference on computer vision and pattern recognition},
  pages={10684--10695},
  year={2022}
}

@article{dalle3,
  title={Improving Image Generation with Better Captions},
  author={Betker, James and Goh, Gabriel and Jing, Li and Brooks, Tim and Wang, Jianfeng and Li, Linjie and Ouyang, Long and Zhuang, Juntang and Lee, Joyce and Guo, Yufei and Manassra, Wesam and Dhariwal, Prafulla and Chu, Casey and Jiao, Yunxin and Ramesh, Aditya},
  year={2023},
  url = "https://cdn.openai.com/papers/dall-e-3.pdf",
}

@article{ye2025echo,
  title={Echo-4o: Harnessing the power of gpt-4o synthetic images for improved image generation},
  author={Ye, Junyan and Jiang, Dongzhi and Wang, Zihao and Zhu, Leqi and Hu, Zhenghao and Huang, Zilong and He, Jun and Yan, Zhiyuan and Yu, Jinghua and Li, Hongsheng and others},
  journal={arXiv preprint arXiv:2508.09987},
  year={2025}
}

@article{bagel,
  title={Emerging properties in unified multimodal pretraining},
  author={Deng, Chaorui and Zhu, Deyao and Li, Kunchang and Gou, Chenhui and Li, Feng and Wang, Zeyu and Zhong, Shu and Yu, Weihao and Nie, Xiaonan and Song, Ziang and others},
  journal={arXiv preprint arXiv:2505.14683},
  year={2025}
}

@article{emu,
  author       = {Quan Sun and
                  Qiying Yu and
                  Yufeng Cui and
                  Fan Zhang and
                  Xiaosong Zhang and
                  Yueze Wang and
                  Hongcheng Gao and
                  Jingjing Liu and
                  Tiejun Huang and
                  Xinlong Wang},
  title        = {Generative Pretraining in Multimodality},
  journal      = {arXiv: 2307.05222},
  year         = {2023},
}

@article{emu2,
  author       = {Quan Sun and
                  Yufeng Cui and
                  Xiaosong Zhang and
                  Fan Zhang and
                  Qiying Yu and
                  Zhengxiong Luo and
                  Yueze Wang and
                  Yongming Rao and
                  Jingjing Liu and
                  Tiejun Huang and
                  Xinlong Wang},
  title        = {Generative Multimodal Models are In-Context Learners},
  journal      = {arXiv: 2312.13286},
  year         = {2023}
}

@inproceedings{xiao2025omnigen,
  title={Omnigen: Unified image generation},
  author={Xiao, Shitao and Wang, Yueze and Zhou, Junjie and Yuan, Huaying and Xing, Xingrun and Yan, Ruiran and Li, Chaofan and Wang, Shuting and Huang, Tiejun and Liu, Zheng},
  booktitle={Proceedings of the Computer Vision and Pattern Recognition Conference},
  pages={13294--13304},
  year={2025}
}

@article{wu2024vila,
  title={Vila-u: a unified foundation model integrating visual understanding and generation},
  author={Wu, Yecheng and Zhang, Zhuoyang and Chen, Junyu and Tang, Haotian and Li, Dacheng and Fang, Yunhao and Zhu, Ligeng and Xie, Enze and Yin, Hongxu and Yi, Li and others},
  journal={arXiv preprint arXiv:2409.04429},
  year={2024}
}

@misc{o1,
  author       = {OpenAI},
  title        = {Introducing openai o1, 2024.},
  year         = {2024},
  url          = {https://openai.com/o1/},
}

@article{guo2025sciverse,
  title={Sciverse: Unveiling the knowledge comprehension and visual reasoning of lmms on multi-modal scientific problems},
  author={Guo, Ziyu and Zhang, Ray and Chen, Hao and Gao, Jialin and Jiang, Dongzhi and Wang, Jiaze and Heng, Pheng-Ann},
  journal={arXiv preprint arXiv:2503.10627},
  year={2025}
}

@article{yue2023mmmu,
  title={MMMU: A Massive Multi-discipline Multimodal Understanding and Reasoning Benchmark for Expert AGI},
  author={Xiang Yue and Yuansheng Ni and Kai Zhang and Tianyu Zheng and Ruoqi Liu and Ge Zhang and Samuel Stevens and Dongfu Jiang and Weiming Ren and Yuxuan Sun and Cong Wei and Botao Yu and Ruibin Yuan and Renliang Sun and Ming Yin and Boyuan Zheng and Zhenzhu Yang and Yibo Liu and Wenhao Huang and Huan Sun and Yu Su and Wenhu Chen},
  journal={arXiv preprint arXiv:2311.16502},
  year={2023},
}

@article{zheng2025deepeyes,
  title={DeepEyes: Incentivizing" Thinking with Images" via Reinforcement Learning},
  author={Zheng, Ziwei and Yang, Michael and Hong, Jack and Zhao, Chenxiao and Xu, Guohai and Yang, Le and Shen, Chao and Yu, Xing},
  journal={arXiv preprint arXiv:2505.14362},
  year={2025}
}

@article{su2025openthinkimg,
  title={Openthinkimg: Learning to think with images via visual tool reinforcement learning},
  author={Su, Zhaochen and Li, Linjie and Song, Mingyang and Hao, Yunzhuo and Yang, Zhengyuan and Zhang, Jun and Chen, Guanjie and Gu, Jiawei and Li, Juntao and Qu, Xiaoye and others},
  journal={arXiv preprint arXiv:2505.08617},
  year={2025}
}

@inproceedings{gao2025interleaved,
  title={Interleaved-modal chain-of-thought},
  author={Gao, Jun and Li, Yongqi and Cao, Ziqiang and Li, Wenjie},
  booktitle={Proceedings of the Computer Vision and Pattern Recognition Conference},
  pages={19520--19529},
  year={2025}
}

@article{shi2025mathcanvas,
  title={MathCanvas: Intrinsic Visual Chain-of-Thought for Multimodal Mathematical Reasoning},
  author={Shi, Weikang and Yu, Aldrich and Fang, Rongyao and Ren, Houxing and Wang, Ke and Zhou, Aojun and Tian, Changyao and Fu, Xinyu and Hu, Yuxuan and Lu, Zimu and others},
  journal={arXiv preprint arXiv:2510.14958},
  year={2025}
}

@article{xu2025visual,
  title={Visual Planning: Let's Think Only with Images},
  author={Xu, Yi and Li, Chengzu and Zhou, Han and Wan, Xingchen and Zhang, Caiqi and Korhonen, Anna and Vuli{\'c}, Ivan},
  journal={arXiv preprint arXiv:2505.11409},
  year={2025}
}

@article{rafailov2023direct,
  title={Direct preference optimization: Your language model is secretly a reward model},
  author={Rafailov, Rafael and Sharma, Archit and Mitchell, Eric and Manning, Christopher D and Ermon, Stefano and Finn, Chelsea},
  journal={Advances in neural information processing systems},
  volume={36},
  pages={53728--53741},
  year={2023}
}

@article{jiang2025cot,
  title={T2i-r1: Reinforcing image generation with collaborative semantic-level and token-level cot},
  author={Jiang, Dongzhi and Guo, Ziyu and Zhang, Renrui and Zong, Zhuofan and Li, Hao and Zhuo, Le and Yan, Shilin and Heng, Pheng-Ann and Li, Hongsheng},
  journal={arXiv preprint arXiv:2505.00703},
  year={2025}
}

@article{zhang2025reasongen,
  title={ReasonGen-R1: CoT for Autoregressive Image generation models through SFT and RL},
  author={Zhang, Yu and Li, Yunqi and Yang, Yifan and Wang, Rui and Yang, Yuqing and Qi, Dai and Bao, Jianmin and Chen, Dongdong and Luo, Chong and Qiu, Lili},
  journal={arXiv preprint arXiv:2505.24875},
  year={2025}
}

@article{pan2025focusdiff,
  title={FocusDiff: Advancing Fine-Grained Text-Image Alignment for Autoregressive Visual Generation through RL},
  author={Pan, Kaihang and Bu, Wendong and Wu, Yuruo and Wu, Yang and Shen, Kai and Li, Yunfei and Zhao, Hang and Li, Juncheng and Tang, Siliang and Zhuang, Yueting},
  journal={arXiv preprint arXiv:2506.05501},
  year={2025}
}

@article{li2024llava-inter,
  title={LLaVA-NeXT-Interleave: Tackling Multi-image, Video, and 3D in Large Multimodal Models},
  author={Li, Feng and Zhang, Renrui and Zhang, Hao and Zhang, Yuanhan and Li, Bo and Li, Wei and Ma, Zejun and Li, Chunyuan},
  journal={arXiv preprint arXiv:2407.07895},
  year={2024}
}

@misc{liu2023improvedllava,
      author={Liu, Haotian and Li, Chunyuan and Li, Yuheng and Lee, Yong Jae},
      title={Improved Baselines with Visual Instruction Tuning}, 
      publisher={arXiv:2310.03744},
      year={2023}
}

@article{wang2024qwen2,
  title={Qwen2-vl: Enhancing vision-language model's perception of the world at any resolution},
  author={Wang, Peng and Bai, Shuai and Tan, Sinan and Wang, Shijie and Fan, Zhihao and Bai, Jinze and Chen, Keqin and Liu, Xuejing and Wang, Jialin and Ge, Wenbin and others},
  journal={arXiv preprint arXiv:2409.12191},
  year={2024}
}

@misc{qwen3technicalreport,
      title={Qwen3 Technical Report}, 
      author={Qwen Team},
      year={2025},
      eprint={2505.09388},
      archivePrefix={arXiv},
      primaryClass={cs.CL},
      url={https://arxiv.org/abs/2505.09388}, 
}

@article{wu2025omnigen2,
  title={OmniGen2: Exploration to Advanced Multimodal Generation},
  author={Wu, Chenyuan and Zheng, Pengfei and Yan, Ruiran and Xiao, Shitao and Luo, Xin and Wang, Yueze and Li, Wanli and Jiang, Xiyan and Liu, Yexin and Zhou, Junjie and others},
  journal={arXiv preprint arXiv:2506.18871},
  year={2025}
}

@article{lin2025uniworld,
  title={Uniworld: High-resolution semantic encoders for unified visual understanding and generation},
  author={Lin, Bin and Li, Zongjian and Cheng, Xinhua and Niu, Yuwei and Ye, Yang and He, Xianyi and Yuan, Shenghai and Yu, Wangbo and Wang, Shaodong and Ge, Yunyang and others},
  journal={arXiv preprint arXiv:2506.03147},
  year={2025}
}

@article{xin2025lumina,
  title={Lumina-dimoo: An omni diffusion large language model for multi-modal generation and understanding},
  author={Xin, Yi and Qin, Qi and Luo, Siqi and Zhu, Kaiwen and Yan, Juncheng and Tai, Yan and Lei, Jiayi and Cao, Yuewen and Wang, Keqi and Wang, Yibin and others},
  journal={arXiv preprint arXiv:2510.06308},
  year={2025}
}

@article{kojima2022large,
  title={Large language models are zero-shot reasoners},
  author={Kojima, Takeshi and Gu, Shixiang Shane and Reid, Machel and Matsuo, Yutaka and Iwasawa, Yusuke},
  journal={Advances in neural information processing systems},
  volume={35},
  pages={22199--22213},
  year={2022}
}

@article{meng2025mm,
  title={MM-Eureka: Exploring Visual Aha Moment with Rule-based Large-scale Reinforcement Learning},
  author={Meng, Fanqing and Du, Lingxiao and Liu, Zongkai and Zhou, Zhixiang and Lu, Quanfeng and Fu, Daocheng and Shi, Botian and Wang, Wenhai and He, Junjun and Zhang, Kaipeng and others},
  journal={arXiv preprint arXiv:2503.07365},
  year={2025}
}

@article{amini2019mathqa,
  title={Mathqa: Towards interpretable math word problem solving with operation-based formalisms},
  author={Amini, Aida and Gabriel, Saadia and Lin, Peter and Koncel-Kedziorski, Rik and Choi, Yejin and Hajishirzi, Hannaneh},
  journal={arXiv preprint arXiv:1905.13319},
  year={2019}
}

@article{hendrycksmath2021,
  title={Measuring Mathematical Problem Solving With the MATH Dataset},
  author={Dan Hendrycks and Collin Burns and Saurav Kadavath and Akul Arora and Steven Basart and Eric Tang and Dawn Song and Jacob Steinhardt},
  journal={NeurIPS},
  year={2021}
}

@article{tschannen2025siglip,
  title={Siglip 2: Multilingual vision-language encoders with improved semantic understanding, localization, and dense features},
  author={Tschannen, Michael and Gritsenko, Alexey and Wang, Xiao and Naeem, Muhammad Ferjad and Alabdulmohsin, Ibrahim and Parthasarathy, Nikhil and Evans, Talfan and Beyer, Lucas and Xia, Ye and Mustafa, Basil and others},
  journal={arXiv preprint arXiv:2502.14786},
  year={2025}
}

@article{kingma2013auto,
  title={Auto-encoding variational bayes},
  author={Kingma, Diederik P and Welling, Max},
  journal={arXiv preprint arXiv:1312.6114},
  year={2013}
}

@article{vaswani2017attention,
  title={Attention is all you need},
  author={Vaswani, Ashish and Shazeer, Noam and Parmar, Niki and Uszkoreit, Jakob and Jones, Llion and Gomez, Aidan N and Kaiser, {\L}ukasz and Polosukhin, Illia},
  journal={Advances in neural information processing systems},
  volume={30},
  year={2017}
}

@inproceedings{esser2024scaling,
  title={Scaling rectified flow transformers for high-resolution image synthesis},
  author={Esser, Patrick and Kulal, Sumith and Blattmann, Andreas and Entezari, Rahim and M{\"u}ller, Jonas and Saini, Harry and Levi, Yam and Lorenz, Dominik and Sauer, Axel and Boesel, Frederic and others},
  booktitle={Forty-first international conference on machine learning},
  year={2024}
}

@article{lipman2022flow,
  title={Flow matching for generative modeling},
  author={Lipman, Yaron and Chen, Ricky TQ and Ben-Hamu, Heli and Nickel, Maximilian and Le, Matt},
  journal={arXiv preprint arXiv:2210.02747},
  year={2022}
}

@article{liu2022flow,
  title={Flow straight and fast: Learning to generate and transfer data with rectified flow},
  author={Liu, Xingchao and Gong, Chengyue and Liu, Qiang},
  journal={arXiv preprint arXiv:2209.03003},
  year={2022}
}

@article{pathiraja2025refedit,
  title={RefEdit: A Benchmark and Method for Improving Instruction-based Image Editing Model on Referring Expressions},
  author={Pathiraja, Bimsara and Patel, Maitreya and Singh, Shivam and Yang, Yezhou and Baral, Chitta},
  journal={arXiv preprint arXiv:2506.03448},
  year={2025}
}

@misc{openai2024gpt4o,
  author    = {OpenAI},
  title     = {Hello GPT-4o},
  howpublished = {\url{https://openai.com/index/hello-gpt-4o/}},
  year      = {2024}
}

@misc{labs2025flux1kontextflowmatching,
      title={FLUX.1 Kontext: Flow Matching for In-Context Image Generation and Editing in Latent Space},
      author={Black Forest Labs and Stephen Batifol and Andreas Blattmann and Frederic Boesel and Saksham Consul and Cyril Diagne and Tim Dockhorn and Jack English and Zion English and Patrick Esser and Sumith Kulal and Kyle Lacey and Yam Levi and Cheng Li and Dominik Lorenz and Jonas Müller and Dustin Podell and Robin Rombach and Harry Saini and Axel Sauer and Luke Smith},
      year={2025},
      eprint={2506.15742},
      archivePrefix={arXiv},
      primaryClass={cs.GR},
      url={https://arxiv.org/abs/2506.15742},
}

@inproceedings{samuel2024generating,
  title={Generating images of rare concepts using pre-trained diffusion models},
  author={Samuel, Dvir and Ben-Ari, Rami and Raviv, Simon and Darshan, Nir and Chechik, Gal},
  booktitle={Proceedings of the AAAI Conference on Artificial Intelligence},
  volume={38},
  number={5},
  pages={4695--4703},
  year={2024}
}

@article{chen2025failureatlas,
  title={FailureAtlas: Mapping the Failure Landscape of T2I Models via Active Exploration},
  author={Chen, Muxi and Zhang, Zhaohua and Zhao, Chenchen and Chen, Mingyang and Jiang, Wenyu and Jiang, Tianwen and Zhuo, Jianhuan and Tang, Yu and Xiao, Qiuyong and Zhang, Jihong and others},
  journal={arXiv preprint arXiv:2509.21995},
  year={2025}
}

@article{liu2023grounding,
  title={Grounding dino: Marrying dino with grounded pre-training for open-set object detection},
  author={Liu, Shilong and Zeng, Zhaoyang and Ren, Tianhe and Li, Feng and Zhang, Hao and Yang, Jie and Li, Chunyuan and Yang, Jianwei and Su, Hang and Zhu, Jun and others},
  journal={arXiv preprint arXiv:2303.05499},
  year={2023}
}

@article{ren2024grounded,
  title={Grounded sam: Assembling open-world models for diverse visual tasks},
  author={Ren, Tianhe and Liu, Shilong and Zeng, Ailing and Lin, Jing and Li, Kunchang and Cao, He and Chen, Jiayu and Huang, Xinyu and Chen, Yukang and Yan, Feng and others},
  journal={arXiv preprint arXiv:2401.14159},
  year={2024}
}

@article{sd3medium2024,
  title={Stable Diffusion 3 Medium},
  author={{Stability AI}},
  journal={arXiv preprint arXiv:2402.13753},
  year={2024},
  note={Model released May 2024}
}

@misc{llamagen2024,
  title={{LlamaGen}: Autoregressive Image Generation with Llama-based Transformers},
  author={{Meta AI}},
  howpublished = {\url{https://github.com/facebookresearch/llamagen}},
  year = {2024},
  note = {Code and model released June 2024}
}

@article{podell2023sdxl,
  title={{SDXL}: Improving Latent Diffusion Models for High-Resolution Image Synthesis},
  author={Podell, Dustin and Marks, Tom and N{\o}lle, Spencer and others},
  journal={arXiv preprint arXiv:2307.01952},
  year={2023},
  url={https://arxiv.org/abs/2307.01952}
}

@article{zhang2025generative,
  title={Generative Universal Verifier as Multimodal Meta-Reasoner},
  author={Zhang, Xinchen and Zhang, Xiaoying and Wu, Youbin and Cao, Yanbin and Zhang, Renrui and Chu, Ruihang and Yang, Ling and Yang, Yujiu},
  journal={arXiv preprint arXiv:2510.13804},
  year={2025}
}

@inproceedings{zhuo2025reflection,
  title={From reflection to perfection: Scaling inference-time optimization for text-to-image diffusion models via reflection tuning},
  author={Zhuo, Le and Zhao, Liangbing and Paul, Sayak and Liao, Yue and Zhang, Renrui and Xin, Yi and Gao, Peng and Elhoseiny, Mohamed and Li, Hongsheng},
  booktitle={Proceedings of the IEEE/CVF International Conference on Computer Vision},
  pages={15329--15339},
  year={2025}
}

@article{qin2025uni,
  title={Uni-cot: Towards unified chain-of-thought reasoning across text and vision},
  author={Qin, Luozheng and Gong, Jia and Sun, Yuqing and Li, Tianjiao and Yang, Mengping and Yang, Xiaomeng and Qu, Chao and Tan, Zhiyu and Li, Hao},
  journal={arXiv preprint arXiv:2508.05606},
  year={2025}
}

@article{zong2024mova,
  title={Mova: Adapting mixture of vision experts to multimodal context},
  author={Zong, Zhuofan and Ma, Bingqi and Shen, Dazhong and Song, Guanglu and Shao, Hao and Jiang, Dongzhi and Li, Hongsheng and Liu, Yu},
  journal={arXiv preprint arXiv:2404.13046},
  year={2024}
}

@article{jiang2024mmsearch,
  title={Mmsearch: Benchmarking the potential of large models as multi-modal search engines},
  author={Jiang, Dongzhi and Zhang, Renrui and Guo, Ziyu and Wu, Yanmin and Lei, Jiayi and Qiu, Pengshuo and Lu, Pan and Chen, Zehui and Song, Guanglu and Gao, Peng and others},
  journal={arXiv preprint arXiv:2409.12959},
  year={2024}
}

@article{ye2025blink,
  title={BLINK-Twice: You see, but do you observe? A Reasoning Benchmark on Visual Perception},
  author={Ye, Junyan and Jiang, Dongzhi and He, Jun and Zhou, Baichuan and Huang, Zilong and Yan, Zhiyuan and Li, Hongsheng and He, Conghui and Li, Weijia},
  journal={arXiv preprint arXiv:2510.09361},
  year={2025}
}

@article{guo2025thinking,
  title={Thinking-while-Generating: Interleaving Textual Reasoning throughout Visual Generation},
  author={Guo, Ziyu and Zhang, Renrui and Li, Hongyu and Zhang, Manyuan and Chen, Xinyan and Wang, Sifan and Feng, Yan and Pei, Peng and Heng, Pheng-Ann},
  journal={arXiv preprint arXiv:2511.16671},
  year={2025}
}

@article{fang2025got,
  title={Got: Unleashing reasoning capability of multimodal large language model for visual generation and editing},
  author={Fang, Rongyao and Duan, Chengqi and Wang, Kun and Huang, Linjiang and Li, Hao and Yan, Shilin and Tian, Hao and Zeng, Xingyu and Zhao, Rui and Dai, Jifeng and others},
  journal={arXiv preprint arXiv:2503.10639},
  year={2025}
}
}

\clearpage

\appendix

\section*{Appendix Overview}
\begin{itemize}
    \item Section~\ref{sup:related_work}: Related work.
    \item Section~\ref{sup:dracocfg}: More Details of \draco{}-CFG.
    \item Section~\ref{sup:dataset}: More Dataset Details.
    \item Section~\ref{sup:exp}: More Experiment Details.
    \item Section~\ref{sup:limit}: Limitations and Future Work.
    \item Section~\ref{sup:qualitative}: Qualitative Examples.
    
\end{itemize}

\section{Related Work}
\label{sup:related_work}
\paragraph{Unified Multimodal Large Language Models}
Many prior works have focused on empowering a single model with both understanding and generation capabilities. Although most methods are built upon MLLMs~\cite{wang2024qwen2, liu2023improvedllava, li2024llava-inter,zong2024mova}, current mainstream approaches have diverged into different streams. 
One direction employs an external diffusion model for image generation~\cite{dong2023dreamllm, emu, emu2, xiao2025omnigen}. For example, OmniGen2~\cite{wu2025omnigen2} uses a diffusion transformer to accept high-level conditions from the MLLM to produce images. 
Another direction adopts the image generation process into the auto-regressive paradigm~\cite{wu2024vila, wu2024janus, chen2025janus}. \citep{wu2024janus} employs CLIP for vision understanding and VQVAE to discretize images for auto-regressive generation. 
Recently, some works have attempted to merge auto-regressive generation and diffusion into a single framework~\cite{zhou2024transfusion, bagel}. Transfusion employs a single transformer model to generate text in an auto-regressive manner and images through iterative denoising. Extending this method further, Bagel employs a mixture of transformer architectures to separately process text tokens and VAE tokens of images.
\paragraph{Reasoning with Images}
With the introduction of OpenAI's o1~\cite{o1}, reasoning with images has been widely explored for image understanding tasks~\cite{Lu2023MathVistaEM,yue2023mmmu,jiang2025mme,zhang2024mathverse,guo2025sciverse, ye2025blink, jiang2024mmsearch}. These methods have attempted to incorporate more information from images into the reasoning process, including the use of external tools~\cite{zheng2025deepeyes, su2025openthinkimg}, the reuse of image tokens from the vision encoder~\cite{gao2025interleaved,chen2025mint}, and the generation of auxiliary images~\cite{xu2025visual, shi2025mathcanvas}.
More recently, the field of visual generation has also adopted this paradigm~\cite{guo2025can, jiang2025cot, gu2025improving, zhang2025reasongen, pan2025focusdiff,guo2025thinking, fang2025got}. Image-Gen-CoT~\cite{guo2025can} evaluates the effectiveness of direct preference optimization~\cite{rafailov2023direct} for image generation and proposes PARM as a reward model for autoregressive generation in test-time scaling. Later, T2I-R1~\cite{jiang2025cot} proposes generating a semantic-level CoT to analyze the given prompt and design the image. The semantic-level CoT is then fed back into the model to generate the image. 
Recently, several methods have proposed reflecting on generated images for improved image generation~\cite{zhuo2025reflection, zhang2025generative, qin2025uni}. \cite{zhuo2025reflection} utilizes a text-to-image diffusion model to generate the image, then a verifier produces reflection and a refined prompt instead of a correction method, and then the generator generates a new image. This method does not sufficiently connect the first generated image with the subsequently generated one, and therefore does not adequately address the difficulty of generating an image in one pass. Concurrently, \cite{zhang2025generative} and \cite{qin2025uni} propose editing the first generated image instead of creating an entirely new one. However, there are several key differences between \draco{} and these methods:
(1) These methods follow a post-reflection strategy, rather than the pre-planning strategy adopted by \draco{}. Therefore, they need to generate the image at the same resolution as the final output and then edit it, incurring much extra cost. 
(2) Additionally, these methods strictly utilize the image editing setting to edit the first generated image, such as restricting the background to remain exactly the same, thereby limiting the editing content.
(3) In our pilot study, we find that the inherent editing capability is insufficient to handle various correction scenarios. Therefore, these methods that rely solely on editing capability cannot sufficiently meet the requirements for correcting errors.

\section{More Details of \draco{}-CFG}
\label{sup:dracocfg}
\subsection{Original CFG of Bagel}
Bagel~\cite{bagel} does not directly support \draco{}, so no CFG can be directly applied to \draco{}. However, the task that most closely resembles \draco{} is "thinking before editing." In this scenario, Bagel takes as input: the image to edit (with both ViT features $vit$ and VAE features $vae$) and an edit instruction $edit$. Bagel then generates a textual chain-of-thought $think$ to analyze how to perform the editing, followed by generating the edited image. Bagel employs three forms of multimodal context in CFG: $m(\phi, \phi, edit, \phi)$, $m(vit, vae, \phi, \phi)$, and the full condition $m(vit, vae, edit, think)$. The final output is computed through two sequential steps:
\begin{equation}
\begin{split}
    \hat{m}(vit, vae, edit, think)'=m(vit, vae, edit, think) \\ 
    + s_{text} \cdot (m(vit, vae, edit, think)-m(vit, vae,\phi,\phi)), \\ 
\end{split}
\label{eq: ori_txt}
\end{equation}
\begin{equation}
\begin{split}
    \hat{m}(vit, vae, edit, think)=\hat{m}(vit, vae, edit, think)' \\ 
    + s_{\text{img}} \cdot (\hat{m}(vit, vae, edit, think)'-m(\phi,\phi, edit, \phi)). 
\end{split}
\label{eq: ori_img}
\end{equation}

This formulation has two key issues: 
\begin{enumerate}
    \item \textbf{Incomplete decoupling of conditions.} The three input conditions, image, edit instruction, and text CoT, are not fully decoupled in the two multimodal contexts used for CFG. Instead, two conditions are always dropped simultaneously. In $m(\phi, \phi, edit, \phi)$, both the image and CoT are absent, while in $m(vit, vae, \phi, \phi)$, both the edit instruction and CoT are absent. This means the CoT is emphasized twice across these two CFG types. Moreover, emphasizing two different conditions simultaneously may produce unexpected results. 
    \item \textbf{Sequential computation couples the scales.} Unlike \draco{}-CFG, which computes CFG in one round, this approach uses two sequential rounds. This couples the scales $s_{text}$ and $s_{img}$. Formally, let $M \triangleq m(vit, vae, edit, think)$, $M_{\text{text}} \triangleq m(vit, vae,\phi,\phi)$, and $M_{\text{img}} \triangleq m(\phi,\phi, edit, \phi)$ denote the latent vectors from the full model, the text-only model, and the image-only model, respectively.
By substituting $\hat{m}(vit, vae, edit, think)'$ from Equation~\ref{eq: ori_txt} into Equation~\ref{eq: ori_img}, we obtain:
\begin{equation}
\begin{split}
\hat{m} &= (1 + s_{text})(1 + s_{\text{img}}) M \\
&\quad - s_{text}(1 + s_{\text{img}}) M_{text} - s_{\text{img}} M_{img}
\end{split}
\label{eq:final_combined}
\end{equation}
As shown, $M_{text}$ is controlled by both $s_{text}$ and $s_{img}$, meaning the condition from $M_{text}$ may be over-emphasized.
\end{enumerate}

\subsection{System Prompt}
In \draco{}-CFG, we adopt different system prompts for $m(\phi,\phi,\phi)$, $m(\phi,{vit},\phi)$, and $m(p, {vit}, v)$ to ensure more emphasis on the specific condition. We omit the details of the system prompt in the full submission for brevity. In practice, we do not adopt system prompt for $m(\phi,\phi,\phi)$. For $m(\phi,{vit},\phi)$, we want the model to fully construct the draft image, so the prompt is:
\begin{tcolorbox}[breakable, colback=gray!5!white, colframe=gray!75!black, 
title=Draft-Only System Prompt, boxrule=0.5mm, width=\linewidth, arc=3mm, auto outer arc]
You should generate a larger image with the same content as the input image, with better details and clarity.

\end{tcolorbox}

For full condition $m(p, {vit}, v)$, the system prompt is:
\begin{tcolorbox}[breakable, colback=gray!5!white, colframe=gray!75!black, 
title=Full-Condition System Prompt, boxrule=0.5mm, width=\linewidth, arc=3mm, auto outer arc]
You should first generate a scratch image. Then you should analyze whether the scratch image is aligned with the prompt. If not, you should think about how to modify the scratch image to make it aligned with the prompt and then modify the scratch image. The analysis process is enclosed within \texttt{\textless{}think\textgreater{}} and \texttt{\textless{}/think\textgreater{}} tags, i.e., 
\texttt{\textless{}think\textgreater{}} analysis process here  modification process here (optional) \texttt{\textless{}/think\textgreater{}} modified image here (optional).
\end{tcolorbox}

\section{More Dataset Details}
\label{sup:dataset}

\begin{figure*}[tp]
    \centering
    \includegraphics[width=\linewidth]{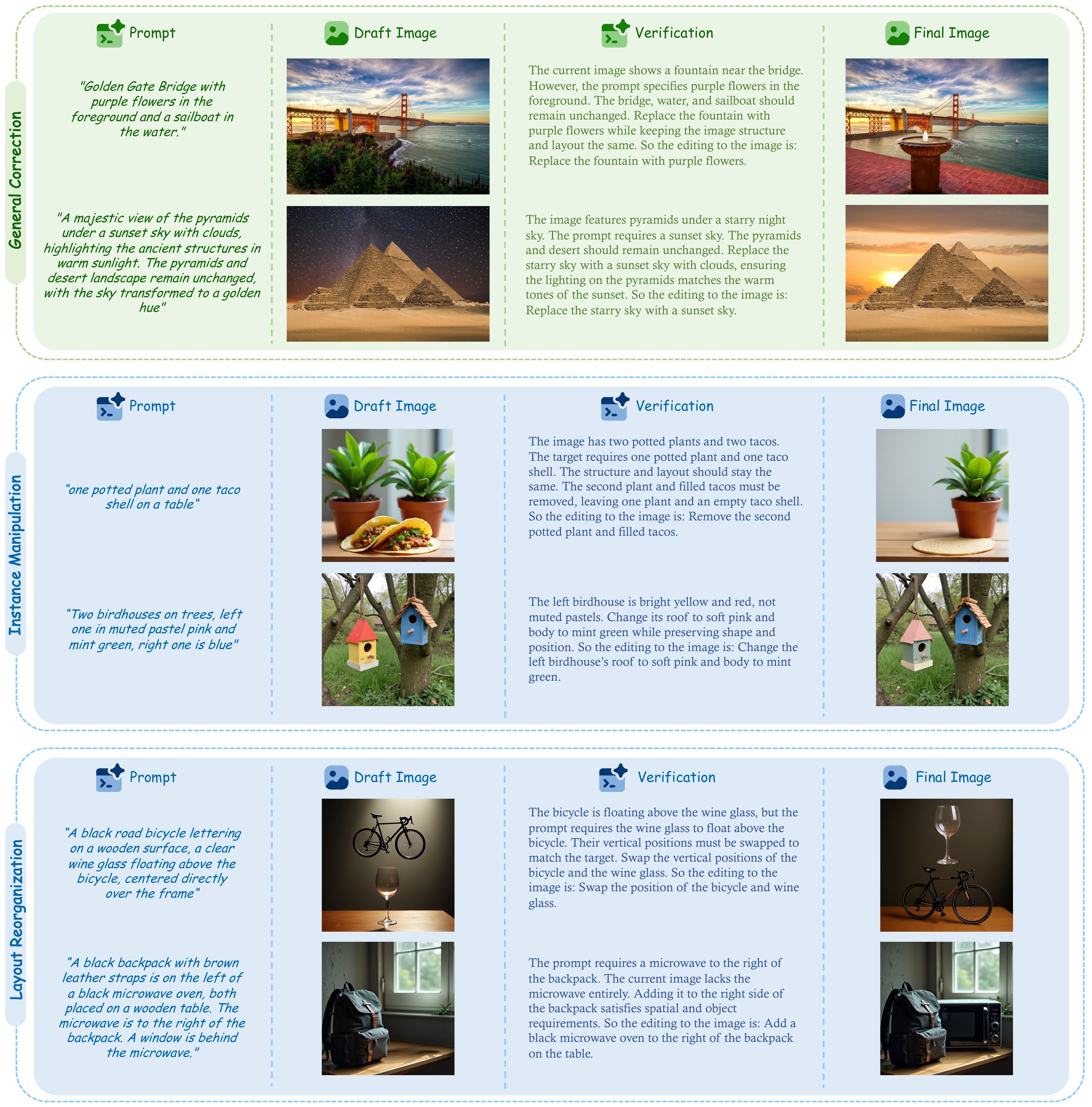}
    \caption{{\bf More Examples of \draco{}-240K.}}
    \label{fig:sup_dataset}
\end{figure*}

We provide additional examples of our proposed \draco{}-240K in Figure~\ref{fig:sup_dataset}.
As shown, the subset for general correction includes various correction scenarios such as object replacement and background modification. 
For instance manipulation, the focus is on handling objects of the same class. The training data includes either adding, removing, or changing attributes for one or multiple instances in the image.
Regarding layout reorganization, we include examples that require swapping the positions of objects, or adding or removing an object at a specific position relative to existing objects.
To guarantee the accuracy of the generated prompts, we conduct cross-validation for prompts in instance manipulation and layout reorganization. The reason is that prompts in instance manipulation contain numeracy information about objects, and prompts in layout reorganization contain spatial information, which is universally acknowledged that current MLLMs, even powerful ones like Qwen3-VL~\cite{qwen3technicalreport}, still struggle to accurately capture. Therefore, we also use GroundingDINO~\cite{liu2023grounding} to detect all the objects present in the prompt and record their positions and numbers. We only retain samples where the detection results align with the prompt.
Besides, the draft images in the dataset are roughly the same size as the final images. We downsample the draft images to a size of 384 $\times$ 384 during training. For the verification, we deliberately add a conclusion about what needs to be changed so that the model can better follow the correction instruction.

\section{More Experiment Details}
\label{sup:exp}
To facilitate Bagel to generate valid small resolution images as a valid draft, we first conduct a text-to-image fine-tuning stage before training for \draco{}. We input various forms of prompts to Bagel to generate 1024 $\times$ 1024 images. Then we resize these images to 384 $\times$ 384 to train the model. We also adopt images generated from GPT-4o~\cite{openai2024gpt4o}. During training, half of the images in trained in 384 $\times$ 384, and half is trained in 1024 $\times$ 1024 to ensure the model's generation capability of both sizes of images. We finetune the model for 14K steps and adopt the EMA weight. We continue training on this weight for \draco{}.

\section{Limitations and Future Work}
\label{sup:limit}
While DraCo demonstrates significant improvements in text-to-image generation through interleaved reasoning, several limitations remain. 
The application of this paradigm to other fields has not been extensively studied. Specifically, the low-resolution draft designed in \draco{} cannot be directly or optimally used in other scenarios, such as videos, 3D assets, or scenes. Clearly, using a low-resolution draft still seems computationally expensive for video generation. The draft for different modalities requires capturing the major difficulty during generation, such as consistency in the video, and then constructing the form of the draft to give a second chance for the most challenging part. These are not clearly studied in this work.
Moreover, the role of humans in improving data curation and the training loop is not well studied. Incorporating humans in this process could help better align the generation and correction methods.

\begin{figure*}
    \centering
    \includegraphics[width=\linewidth]{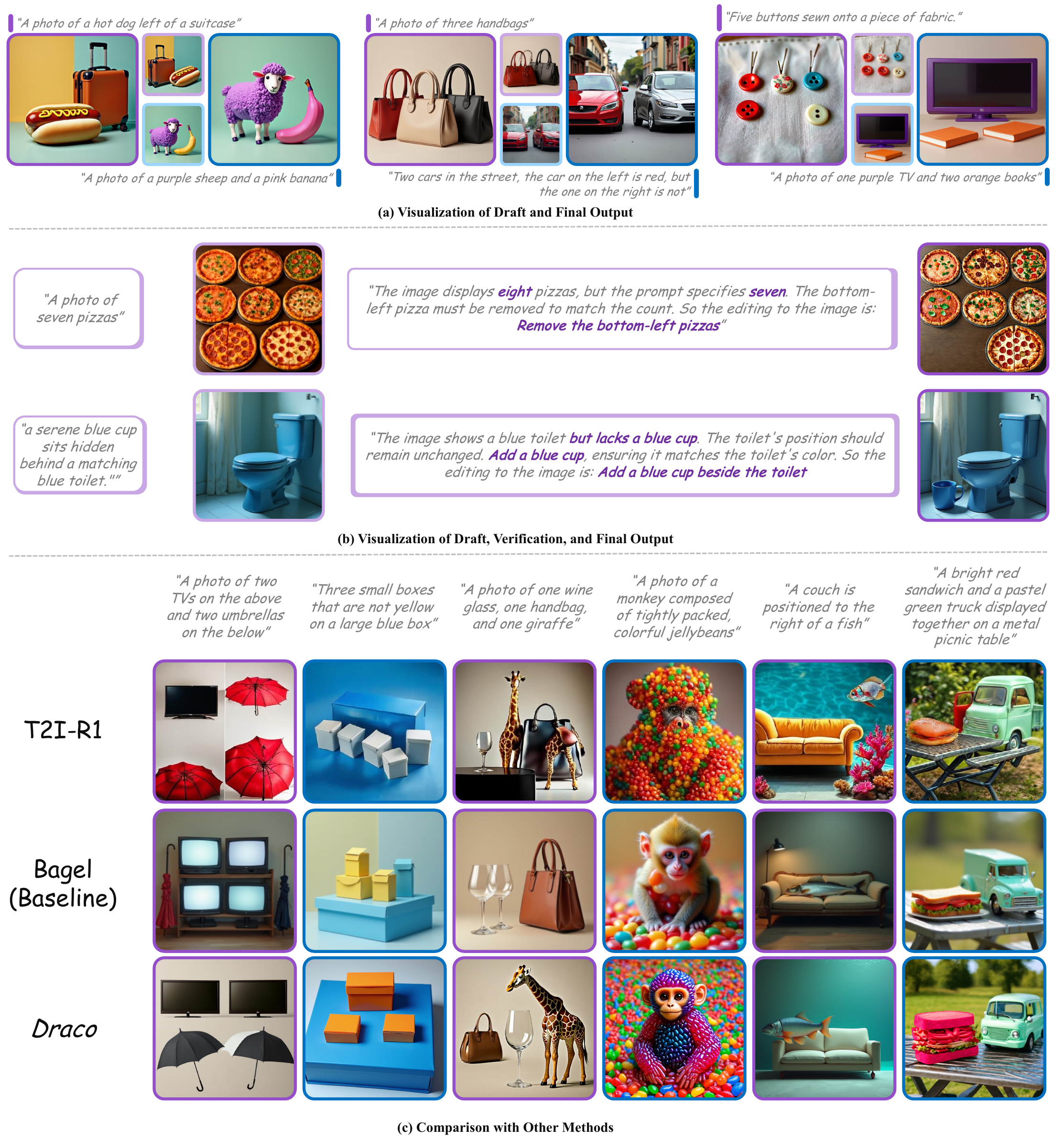}
    \caption{{\bf More Qualitative Examples of \draco{}.}}
    \label{fig:sup_example}
\end{figure*}

\section{Qualitative Examples}
\label{sup:qualitative}
We show more qualitative examples and comparison in Fig.~\ref{fig:sup_example}. 
In Fig.~\ref{fig:sup_example} (a), we showcase the draft image, final image, and its corresponding prompt. The examples include various correction scenarios, including: position correction, rare attribute generation, precise object editing, and numeracy correction. 
We also show more detailed examples including the verification in Fig.~\ref{fig:sup_example} (b). During verification, the model first analyzes the inconsistency between the prompt and the image, then proposes a correction method, and also highlights what should not be changed. Ultimately, the model summarizes the correction in its final outcome.
In Fig.~\ref{fig:sup_example} (c), we compare our method with the CoT-powered method, T2I-R1~\cite{jiang2025cot}, and our baseline model, Bagel~\cite{bagel}. As shown, T2I-R1 tends to generate artifacts, e.g., half of the giraffe and the monkey, or over-saturated images. While Bagel cannot accurately follow the prompt. In contrast, \draco{} produces satisfying results with both high quality and precise alignment of the prompt.

\clearpage

\end{document}